\documentclass{article}

\usepackage{arxiv}

\usepackage[utf8]{inputenc} % allow utf-8 input
\usepackage[T1]{fontenc}    % use 8-bit T1 fonts
\usepackage{hyperref}       % hyperlinks
\usepackage{url}            % simple URL typesetting
\usepackage{booktabs}       % professional-quality tables
\usepackage{amsfonts}       % blackboard math symbols
\usepackage{nicefrac}       % compact symbols for 1/2, etc.
\usepackage{microtype}      % microtypography
\usepackage{lipsum}		% Can be removed after putting your text content
\usepackage{graphicx}
\usepackage{natbib}
\usepackage{doi}
\usepackage{xcolor}
\usepackage{subcaption}
\usepackage{hyperref}
\usepackage{placeins}
\usepackage{float}

\title{ChemAU: Harness the Reasoning of LLMs in Chemical Research with Adaptive Uncertainty Estimation}

\author{
 Xinyi Liu$^1$, Lipeng Ma$^1$, Yixuan Li$^1$, Weidong Yang$^{1,}$\thanks{Corresponding authors}, Qingyuan Zhou$^1$, Jiayi Song$^1$, Shuhao Li$^1$, Ben Fei$^{2,}$\protect\footnotemark[1]\\
  $^1$ Fudan University, $^2$ The Chinese University of Hong Kong\\
  \texttt{24210240236@m.fudan.edu.cn, lpma21@m.fudan.edu.cn, wdyang@fudan.edu.cn, benfei@cuhk.edu.hk}
  }

\date{}

\renewcommand{\shorttitle}{ChemAU}

\hypersetup{
pdftitle={A template for the arxiv style},
pdfsubject={q-bio.NC, q-bio.QM},
pdfauthor={David S.~Hippocampus, Elias D.~Striatum},
pdfkeywords={First keyword, Second keyword, More},
}

\begin{document}
\maketitle

\begin{abstract}
  Large Language Models (LLMs) are widely used across various scenarios due to their exceptional reasoning capabilities and natural language understanding. 
  While LLMs demonstrate strong performance in tasks involving mathematics and coding, their effectiveness diminishes significantly when applied to chemistry-related problems.
  Chemistry problems typically involve long and complex reasoning steps, which contain specific terminology, including specialized symbol systems and complex nomenclature conventions. 
  These characteristics often cause general LLMs to experience hallucinations during the reasoning process due to their lack of specific knowledge. 
  However, existing methods are struggling to effectively leverage chemical expertise and formulas. Moreover, current uncertainty estimation methods, designed to mitigate potential reasoning errors, are unable to precisely identify specific steps or key knowledge. 
  In this work, we propose a novel framework called \textbf{ChemAU}, which incorporates our adaptive uncertainty estimation method that applies different uncertainty values based on the position of reasoning steps within the whole reasoning chain. 
  Leveraging this method, ChemAU identifies gaps in chemistry knowledge and precisely supplements chemical expertise with the specialized domain model, thereby correcting and updating the previously flawed reasoning chain. 
  Our experiments with three popular LLMs across three chemistry datasets demonstrate that ChemAU significantly enhances both reasoning accuracy and uncertainty estimation. 
% Code is available at \url{https://anonymous.4open.science/r/ChemAU-1F72}. 
\end{abstract}

% keywords can be removed
% \keywords{First keyword \and Second keyword \and More}

\section{Introduction}

\label{sec:1}

In recent years, large language models (LLMs) have undergone dramatic advancement and demonstrated remarkable utility across multiple fields~\cite{topsakal2023creating, achiam2023gpt, guo2025deepseek}, such as natural language processing\cite{zubiaga2024natural}, computer vision\cite{sapkota2024synthetic}, legal and medical fields~\cite {li2024legalagentbench, goyal2024healai}. 
Beyond these applications, reasoning techniques such as chain-of-thought (CoT)~\cite{wei2022chain}, self-reflection~\cite{renze2024self} have been developed to significantly enhance the inferential capabilities of LLMs, revealing their substantial potential in scientific domains. 
Notably, LLMs have been applied to assist in various chemistry tasks, including molecular property prediction~\cite{qian2023can} and experimental protocol design~\cite{huang2024crispr}. 
These applications demonstrate that LLMs possess significant potential for supporting chemical research and addressing chemistry-related problems.

Research in chemistry-focused LLMs predominantly follows two main approaches. 
The first approach follows the \textit{pre-train and fine-tune} paradigm, developing domain-specific models from scratch. These models are first pre-trained on specialized chemical data, such as SMILES or SELFIES molecular~\cite{wang2019smiles, honda2019smiles, bagal2021molgpt}, to learn domain-specific features. They are then fine-tuned on task-specific datasets to optimize for objectives like chemical toxicity prediction and drug solubility prediction~\cite{axelrod2022geom}. 
However, unlike general-purpose LLMs, which leverage vast amounts of general text data, these models are typically smaller in scale and rely on limited, high-quality datasets curated through domain expertise~\cite{zhang2025scientific}. This reliance on narrowly focused data restricts their scalability and flexibility, as they are optimized for specific tasks and operate within strictly defined input-output formats.

The second approach leverages general LLMs by instruction-tuning them with domain-specific chemical knowledge~\cite{zhang2024chemllm, li2025chemvlm}. This method enhances the model's expertise in chemistry while retaining its general-purpose capabilities. Unlike traditional small-scale models, these chemistry-specialized LLMs support diverse input formats, flexible task requirements, and coherent dialogue capabilities. However, their large parameter sizes and the need for extensive domain-specific training data significantly increase computational costs and demand substantial resources. Moreover, this approach is typically feasible only for open-source LLMs, limiting broader applicability. 
Another emerging LLM-based approach is Retrieval-Augmented Generation (RAG)~\cite{asai2023self}, which integrates domain-specific knowledge into LLMs through information retrieval mechanisms rather than parameter updates. While RAG reduces the need for extensive fine-tuning, the retrieved knowledge fragments often lack coherence and accuracy, introducing cognitive noise into the reasoning process. This fragmentation can negatively affect the model's performance on complex chemistry tasks, where precise and contextual knowledge is crucial.

To address these challenges, we propose a novel framework that synergizes the powerful reasoning capabilities of general LLMs with the specialized domain knowledge of chemistry-specific models. Drawing inspiration from recent advancements in uncertainty estimation~\cite{liu2025uncertainty, huang2023look}, our framework incorporates a step-by-step uncertainty estimation mechanism. This mechanism dynamically evaluates the confidence of the general LLM at each reasoning step, identifying when to invoke the specialized model for domain-specific knowledge supplementation. By leveraging the complementary strengths of general and specialized models, our framework ensures both accuracy and reliability in tackling complex chemistry problems.

\begin{figure}[t]
    \centering
    \includegraphics[width=\textwidth]{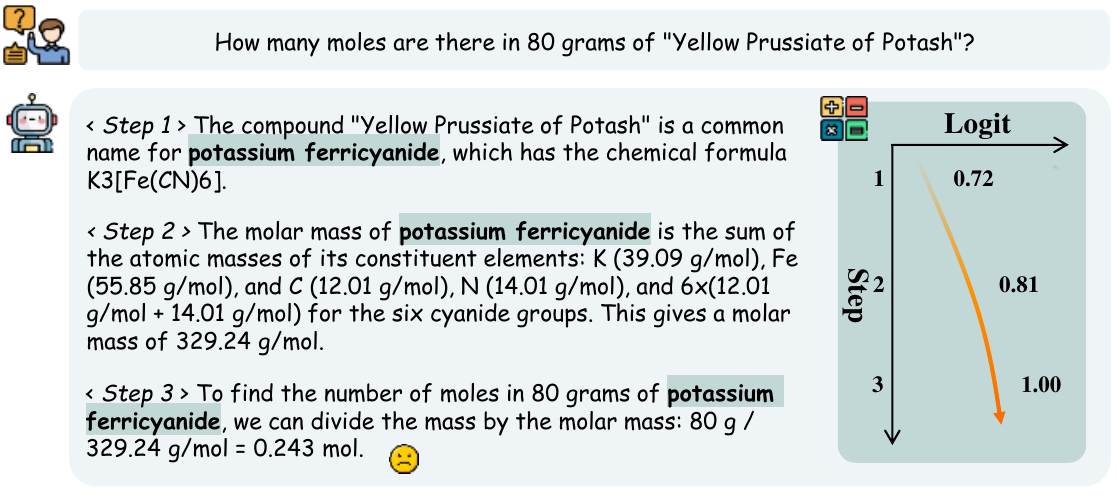}
    \caption{Chemistry-specific tokens exhibit a distinctive pattern during the reasoning process: as reasoning steps increase, their logit values progressively rise. In this chemistry problem, the LLM generates three consecutive reasoning steps, with the chemistry token ``\textbf{potassium ferricyanide}'' appearing in each step. 
    The data reveals that in the first reasoning step, the logit value for this token is \textit{0.72}. In the second step, it increases to \textit{0.81}. And by the third step, the logit value further grows to \textit{1.0}. This phenomenon clearly demonstrates that as the reasoning chain increases, the logit values of specific chemistry tokens show an upward trend. }
    \label{fig:Finding}
\end{figure}

Recent research on uncertainty estimation in LLMs has introduced methods such as token-level probability~\cite{ling2024uncertainty}, self-verbalized~\cite{tang2024evaluation}, and semantic-similarity approaches~\cite{chen2023quantifying}, which typically rely on predefined thresholds to classify responses as reliable or unreliable.
However, in chemistry-specific tasks, we observe a unique phenomenon: the logit values of chemistry-specific tokens, initially low, progressively increase as reasoning advances. This behavior, as illustrated in Figure~\ref{fig:Finding}, arises because LLMs, primarily trained on general corpora, struggle to represent domain-specific symbols and nomenclature effectively. Over time, the model begins to treat these tokens as ``active vocabulary'', artificially inflating their probabilities to maintain thematic consistency. This phenomenon undermines the accuracy of existing uncertainty estimation methods, particularly in domains with unique terminologies like chemistry. To address this issue, we propose a novel step-wise uncertainty estimation method that dynamically adjusts uncertainty values based on the position of each reasoning step within the reasoning chain. This approach allows the model to better identify when domain-specific expertise is needed, facilitating more effective collaboration between general and specialized models.

\begin{figure}[t]
    \centering
    \includegraphics[width=\textwidth]{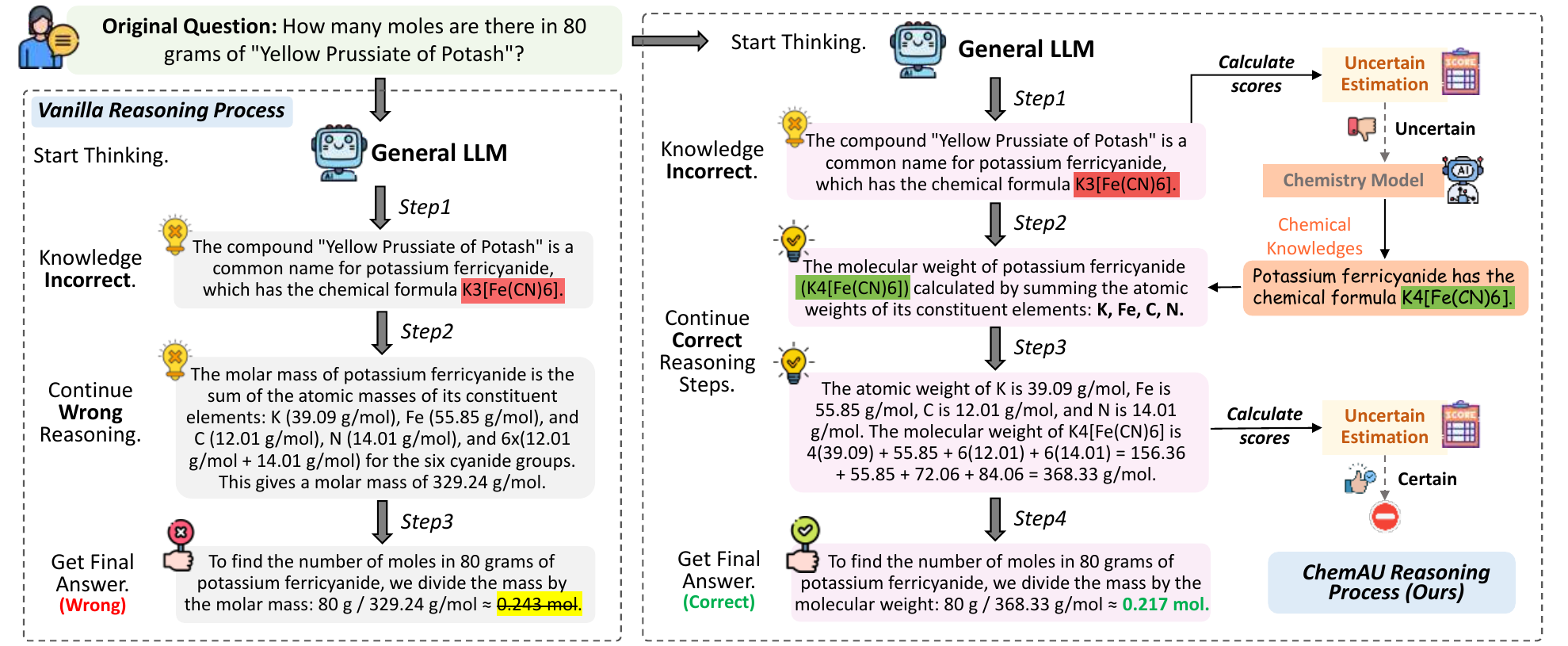}
    \caption{Overview of our proposed LLM reasoning framework with \textbf{A}daptive \textbf{U}ncertainty estimation specifically designed for \textbf{Chem}istry problems (\textbf{ChemAU}). The left side illustrates the reasoning process of a conventional general LLM when solving chemistry problems, which failed to detect the error in the chemical formula ``$K_3[Fe(CN)_6]$'', ultimately resulting in an incorrect inference. The right side demonstrates how our proposed framework successfully identifies the error in \textit{Step1} through an uncertainty estimation mechanism, subsequently redirecting the query to a specialized chemistry domain model to obtain accurate domain knowledge. Specifically, the correct molecular formula ``$K_4[Fe(CN)_6]$''. This accurate information is then reintroduced into the general LLM, ultimately yielding the correct answer. }
    \label{fig:Framework}
\end{figure}

As shown in Figure~\ref{fig:Framework}, we introduce \textbf{ChemAU}, a novel LLM reasoning framework with \textbf{A}daptive \textbf{U}ncertainty estimation specifically designed for \textbf{Chem}istry reasoning tasks. 
ChemAU employs a general LLM to generate a reasoning chain for a given chemistry question and sequentially evaluates the uncertainty of each reasoning step. Steps identified with high uncertainty—often associated with unfamiliar chemistry-specific tokens—trigger the specialized chemistry model to analyze the accuracy of the current step and provide relevant domain knowledge. The provided domain-specific knowledge is then integrated into the reasoning process to guide subsequent steps, ensuring accurate and contextually relevant outputs.

We evaluate the ChemAU framework with three popular general LLMs (\textit{Qwen2.5-7B-Instruct}~\cite{yang2024qwen2}, \textit{LLaMA-3-8B-Instruct}~\cite{grattafiori2024llama}, \textit{DeepSeek-R1-Distill-Qwen-14B}~\cite{guo2025deepseek}) across three distinct chemistry datasets (GPQA~\cite{rein2024gpqa}, MMLU-Pro~\cite{wang2024mmlu}, SuperGPQA~\cite{du2025supergpqa}). Experimental results demonstrate that our proposed framework significantly improves the performance of LLMs on chemistry problems. 
Our contributions can be listed as follows:
\begin{itemize}
    \item We propose an LLM reasoning framework with \textbf{A}daptive \textbf{U}ncertainty estimation specifically designed for \textbf{Chem}istry problems (\textbf{ChemAU}), which combines the powerful reasoning capabilities of general LLMs with the precise domain knowledge of specialized chemistry models. To the best of our knowledge, this is the first framework to introduce a model collaboration strategy for chemistry reasoning tasks. 

    \item We identify a unique phenomenon in chemistry-specific problems where the logit values of domain tokens progressively rise during reasoning, leading to inaccurate uncertainty estimation with existing methods. To address this, we propose a step-wise uncertainty estimation method that dynamically adjusts uncertainty values based on their position within the reasoning chain. 

    \item Extensive experiments across multiple general LLMs and chemistry datasets demonstrate that ChemAU significantly improves performance in chemistry reasoning tasks, highlighting its potential for advancing domain-specific applications. 
\end{itemize}

\section{Related Work}

\label{sec:2}

\textbf{Uncertainty Estimation of LLMs. }
Due to the substantial computational expense of running inference on LLMs, the LLM community has largely moved away from traditional uncertainty estimation methods used for learned models~\cite{balabanov2024uncertainty, shorinwa2024survey}. Instead, researchers have developed less computationally demanding approximate techniques that leverage the distinctive architecture of LLMs to evaluate model uncertainty~\cite{azaria2023internal, yang2024qwen2, huang2023look, zhang2024luq}. These uncertainty estimation approaches for LLMs can be generally divided into two categories: white-box models and black-box models~\cite{fadeeva2023lm}. 

White-box uncertainty estimation methods leverage access to intermediate outputs from the underlying system, such as the probability distributions across generated tokens, to evaluate and quantify the model's uncertainty~\cite{manakul2023selfcheckgpt, fadeeva2024fact}. In contrast, black-box uncertainty estimation methods assess a model's uncertainty solely from its responses to input prompts. These methods typically require evaluating the similarity between multiple responses generated by either a single LLM or an ensemble of LLMs~\cite{liu2019roberta, reimers2019sentence, zhang2019bertscore}. However, the characteristic of black-box methods requiring multiple generations determines that their usage consumes substantial computational resources. Our proposed uncertainty estimation method only requires a single generation for white-box models, greatly improving computational efficiency.

There are mainly two kinds of uncertainty: aleatoric uncertainty and epistemic uncertainty~\cite{hou2023decomposing}. Aleatoric uncertainty occurs in situations where there is inherent randomness and noise in the data-generating process, which means it is inevitable~\cite{min2020ambigqa, kuhn2022clam}. In contrast, epistemic uncertainty can be reduced or eliminated because it occurs due to a lack of knowledge or limited training data. Our proposed framework focuses on addressing the problem of inaccurate responses generated by LLMs due to knowledge deficiencies.

\textbf{Applications of LLMs in Chemistry. }
LLMs have gained impressive success in natural language understanding and complex question reasoning~\cite{hadi2023survey, li2024personal}. Its strong abilities have been applied in various kinds of chemistry related tasks~\cite{jablonka2024leveraging, li2024empowering, boiko2023autonomous}. The current approach for chemistry-specific LLMs, like ChemLLM~\cite{zhang2024chemllm} and ChemDFM~\cite{zhao2024chemdfm}, involves collecting data from research papers and textbooks, pre-training domain knowledge on general LLMs, and then fine-tuning with chemical instructions. These models perform excellently on chemistry-related tasks and demonstrate significant advantages in terms of accuracy and depth of domain knowledge~\cite{huang2024chemeval}. However, precisely because of their heightened focus on chemistry domain knowledge and tasks, they may emphasize factual knowledge and professional task performance over reasoning processes, potentially limiting their reasoning capabilities compared to general LLMs.

\section{Method}

\textbf{Chem}ical Research with \textbf{A}daptive \textbf{U}ncertainty Estimation (\textbf{ChemAU}) aims to enhance the reasoning performance of LLMs in the chemistry domain through uncertainty estimation and precise supplementation of missing knowledge. 
As depicted in Figure~\ref{fig:Framework}, we first input a chemistry problem into the general LLM to generate an initial reasoning chain. 
Subsequently, we perform uncertainty estimation individually for each reasoning step. 
For the step with high uncertainty, we utilize domain-specific models to provide accurate chemistry knowledge related to the concepts mentioned. 
This accurate information, along with the previous correct reasoning steps, is then input to the general LLM, prompting the model to regenerate a more reliable subsequent reasoning chain. 
Our method consists of three components: Adaptive Uncertainty Estimation, Extraction and Supplementation of Chemistry Knowledge in a Single Reasoning Step, and Adjustment of Reasoning Steps. 

\subsection{Adaptive Uncertainty Estimation}

\label{sec:3.1}

For general problems, probability-based methods commonly employ length-normalized scoring for uncertainty measurement~\cite{malinin2020uncertainty}, wherein equal weighting is applied to all tokens in the generated sequence, with these weights being inversely correlated to the total sequence length. To mitigate the impact of probability from function words such as ``the'', ``an'', and ``of'' on uncertainty estimation, several studies have proposed determining tokens' weights based on their semantic contribution to the entire sentence~\cite{bakman2024mars, duan2023shifting}. 
However, models detecting semantic similarity may struggle to discern differences in domain-specific knowledge between sentences due to the lack of chemistry expertise in their pre-training data. 
More importantly, they may incorrectly identify two sentences as semantically equivalent even when their key components are fundamentally different, leading to a misjudgment, as shown in Appendix. 

Another probability-based method uses the maximum value of the negative log probabilities of all tokens as an uncertainty estimation metric~\cite{manakul2023selfcheckgpt,shorinwa2024survey}:
\begin{equation}
    Max(-log\ p) = \mathop{max}\limits_j\ -log\ (p_j),
\end{equation}
which evaluates by identifying the token with the lowest likelihood.  
However, this method does not perform well in chemistry domain. 
Based on our observations, we find a notable phenomenon: when LLMs generate responses to chemistry-specific inquiries, chemistry-specialized tokens typically exhibit markedly low logit values upon initial appearance. However, as the chain of reasoning extends and develops, the frequency of these specialized terms tends to rise, accompanied by a corresponding increase in their logit scores, as illustrated in Figure~\ref{fig:Finding}. 
This phenomenon may stem from the fact that LLMs undergo training mainly on general textual data, while the chemistry domain often incorporates unique symbolic systems and specialized expressions that appear infrequently in common texts, resulting in limited representation learning. As the reasoning process continues, the model progressively recognizes these chemistry-specific terms as contextually relevant tokens, consequently enhancing their prediction probabilities to preserve topical coherence. 

The methods mentioned above demonstrate good performance on general problems, typically employing a predefined threshold where expressions with uncertainty above it are considered potentially erroneous. However, according to our findings, applying a fixed threshold to reasoning steps at different positions within the reasoning chain presents significant limitations when facing chemistry-related problems. 

To address these challenges, we propose a novel uncertainty estimation method specifically designed for LLMs' reasoning in the chemistry domain. This method dynamically assigns distinct uncertainty values to each reasoning step based on its position within the overall reasoning chain: 
\begin{equation}
    \textbf{U}_i(\textbf{R},\ \textbf{P}_i)=\mathop{max}\limits_j -log(p_{ij})\ +\ \alpha(\textbf{L}_\textbf{R}\ -\ i),
\end{equation}
where $U_i$ quantifies the uncertainty of the $i$-th reasoning step, $R$ represents the complete reasoning chain, $\alpha$ represents a predefined constant, $L_R$ represents the number of reasoning steps, and $P_i$ indicates the probabilities of all tokens (denoted as $p_{ij}$ for each token) within the $i$-th step. 
If the uncertainty exceeds a predefined threshold $\theta$, formally expressed as:
$
    \mbox{if}\ U_i(R,\ P_i) > \theta,
$
indicating that this reasoning step exhibits a high likelihood of containing potential errors and requires further processing.

\subsection{Extraction and Supplementation of Chemistry Knowledge in a Single Reasoning Step}

\label{sec:3.2}

For a potentially erroneous step, we decompose it into multiple units of atomic chemistry knowledge for analysis and correction. 
Through this fine-grained decomposition, we can precisely identify and supplement knowledge deficiencies in the chemistry domain that LLMs exhibit during reasoning chain generation. 
Specifically, we construct a chemistry knowledge dataset and use it to perform instruction fine-tuning on \textit{Qwen2.5-1.5B-Instruct}~\cite{yang2024qwen2}, developing a specialized chemistry domain model. 
This model evaluates the accuracy of input chemistry knowledge points. 
When it detects inaccuracies or incompleteness, it promptly provides corresponding precise and comprehensive chemistry knowledge in response. 
By feeding decomposed atomic-level chemistry knowledge points into this chemical-specialized model, its output can accurately remedy knowledge gaps in the general LLM within the chemistry domain, effectively correcting the current reasoning step. 
This strategy, instead of exhaustively verifying all chemistry knowledge, allows us to concentrate computational resources only on the steps that are most likely to contain errors, precisely identifying and supplementing missing knowledge where needed, thereby improving the overall efficiency of reasoning.

\subsection{Adjustment of Reasoning Steps}

Chemistry problems typically exhibit the following characteristics: 
\begin{itemize}
    \item They involve long chains of thought, which require multiple reasoning steps to solve a problem. 

    \item The reasoning steps are tightly interconnected, where each step usually serves as the foundation for the next. 
\end{itemize}

\begin{figure}[t]
    \centering
    \includegraphics[width=\textwidth]{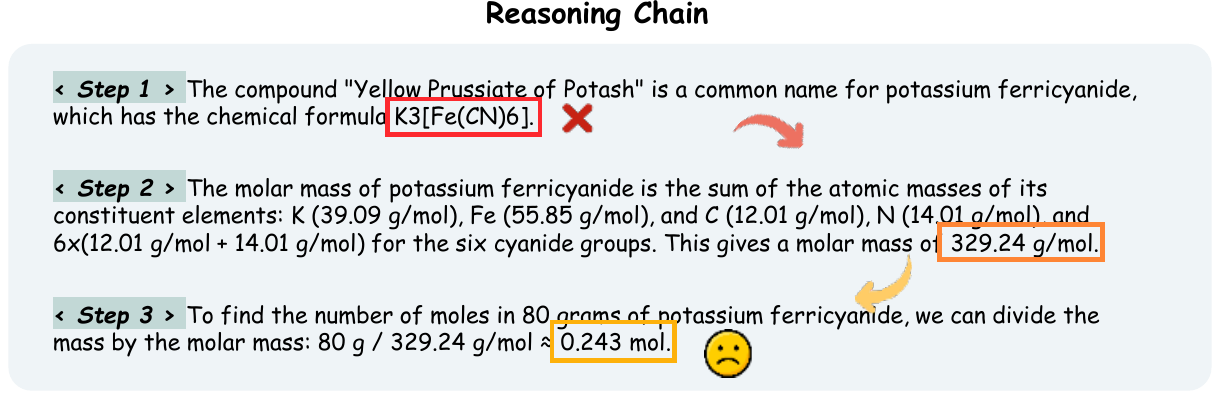}
    \caption{In chemistry problem-solving, reasoning steps are closely interconnected, with each subsequent step typically building upon the results of the previous one. For instance, in the illustrated example, the molar amount ``0.243'' (calculated in < \textit{Step 3} >) is derived from the molar mass ``329.24'' (determined in < \textit{Step 2} >), which in turn is based on the molecular formula ``$K_3[Fe(CN)_6]$'' (identified in < \textit{Step 1} >). Consequently, errors in the first step will propagate through the entire reasoning chain, compromising the accuracy of all subsequent calculations, similar to a domino effect. }
    \label{fig:Process}
\end{figure}

As shown in the example provided in Figure~\ref{fig:Process}, the correctness of the ``\textbf{moles}'' in < \textit{Step 3} > directly depends on the accuracy of ``\textit{molar mass}'' from < \textit{Step 2} >, which bases on the exactness of ``\textit{chemical formula}'' from < \textit{Step 1} >. 
Therefore, the certainty for earlier reasoning steps should be stricter, precisely matching the dynamic uncertainty estimation approach introduced in Section~\ref{sec:3.1} that assigns higher uncertainty values to reasoning steps at earlier positions, rather than applying the same value across all reasoning steps.

For the entire reasoning process, we propose a reasoning framework shown in Figure~\ref{fig:Framework}. 
First, we input a specific chemistry problem into the general LLM to generate an initial reasoning chain. 
Subsequently, we implement a sequential detection strategy throughout the reasoning chain, evaluating each step's uncertainty value in order. When the uncertainty value is low, we consider the step correct and add it to the set of confirmed reasoning steps. 
Once we identify a reasoning step whose uncertainty value exceeds the predetermined threshold, we exclusively focus on that particular step, temporarily setting aside subsequent reasoning processes. 
For this high-uncertainty step, we perform atomic chemistry knowledge extraction, generating corresponding expertise through the chemistry-specific model mentioned in Section~\ref{sec:3.2} to bridge the knowledge gap in the general LLM. 
Then, we reintroduce the newly acquired chemistry knowledge, the original chemistry problem, and the confirmed reasoning steps back into the general LLM to continue generating a new reasoning chain. 
This iterative process continues, forming a complete reasoning procedure that adjusts to the chemistry domain problem.

\label{sec:method}

\section{Experiment}
\label{sec:4}

\subsection{Experimental Settings}
\label{sec:4.1}

\textbf{Models. }
In our experiments, we utilize three different series of open-source LLMs, including \textit{Qwen2.5-7B-Instruct}~\cite{yang2024qwen2}, \textit{LLaMA-3-8B-Instruct}~\cite{grattafiori2024llama}, and \textit{DeepSeek-R1-Distill-Qwen-14B}~\cite{guo2025deepseek}, with model sizes of 7B, 8B, and 14B. Additionally, based on our constructed chemistry domain dataset, we perform instruction fine-tuning on the \textit{Qwen2.5-1.5B-Instruct}\cite{yang2024qwen2} to serve as the domain-specific model for our experiments. 

\textbf{Datasets. }
We utilize three distinct datasets, specifically \textit{GPQA}~\cite{rein2024gpqa}, \textit{MMLU-Pro}~\cite{wang2024mmlu}, and \textit{SuperGPQA}~\cite{du2025supergpqa}, from which we extract chemistry-related questions for our experimental analysis. 

They contain complex chemistry domain problems that require both multi-step reasoning processes and specialized knowledge, creating a strong contrast with commonsense questions and providing an ideal testing scenario for evaluating LLMs' reasoning capabilities in chemistry.

\textbf{Baselines. }
We evaluate all models using Chain-of-Thought (CoT) prompting to encourage step-by-step reasoning, applying identical prompting templates across all experiments to ensure fair comparison. Templates are shown in Appendix. 
For each dataset, we measure the answer accuracy as our primary evaluation metric. 

\begin{figure}[htbp]
    \centering
    \begin{subfigure}[b]{0.31\textwidth}
        \centering
        \includegraphics[width=\textwidth]{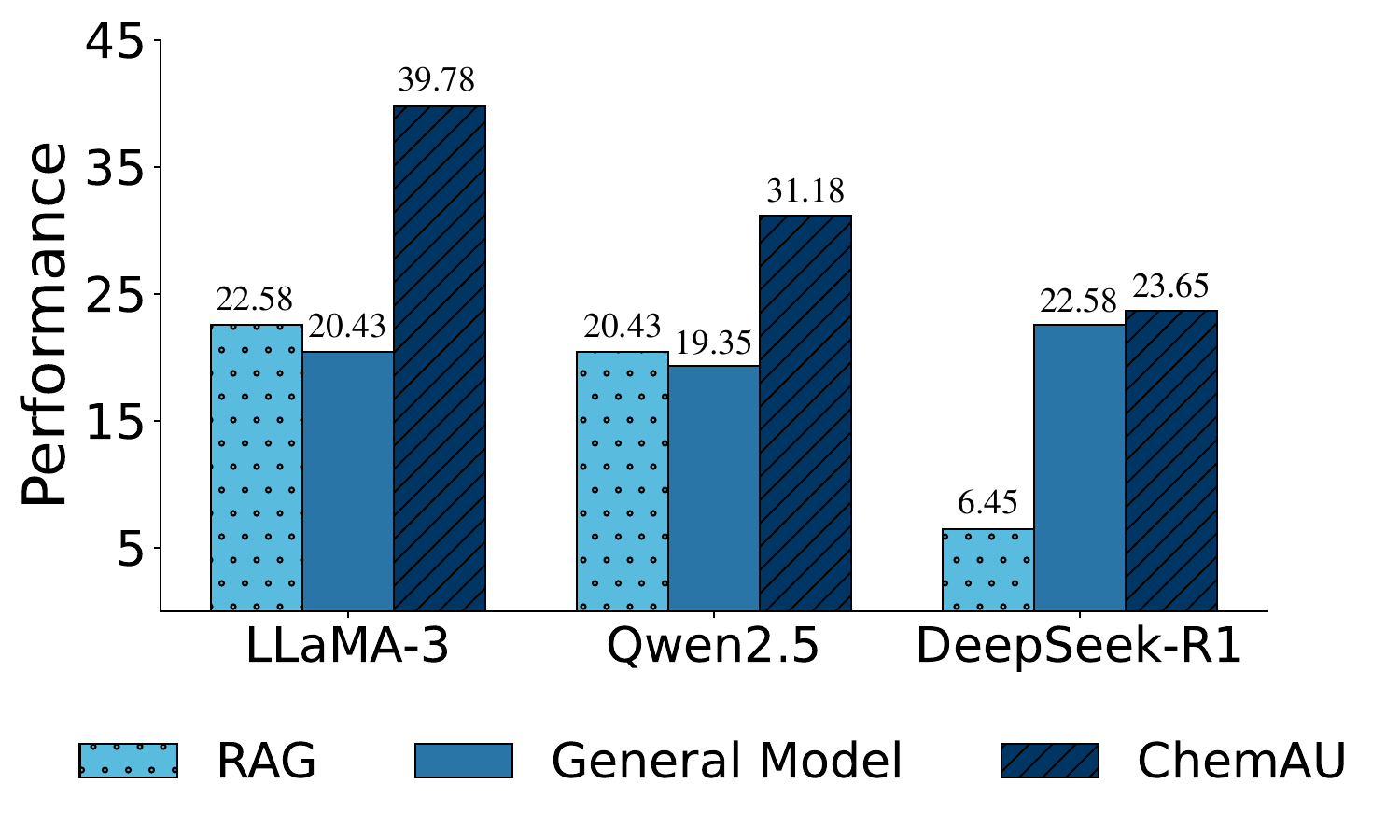}
        \caption{GPQA}
        \label{fig:reasoning_performance_comparison_gpqa}
    \end{subfigure}
    \hfill
    \begin{subfigure}[b]{0.31\textwidth}
        \centering
        \includegraphics[width=\textwidth]{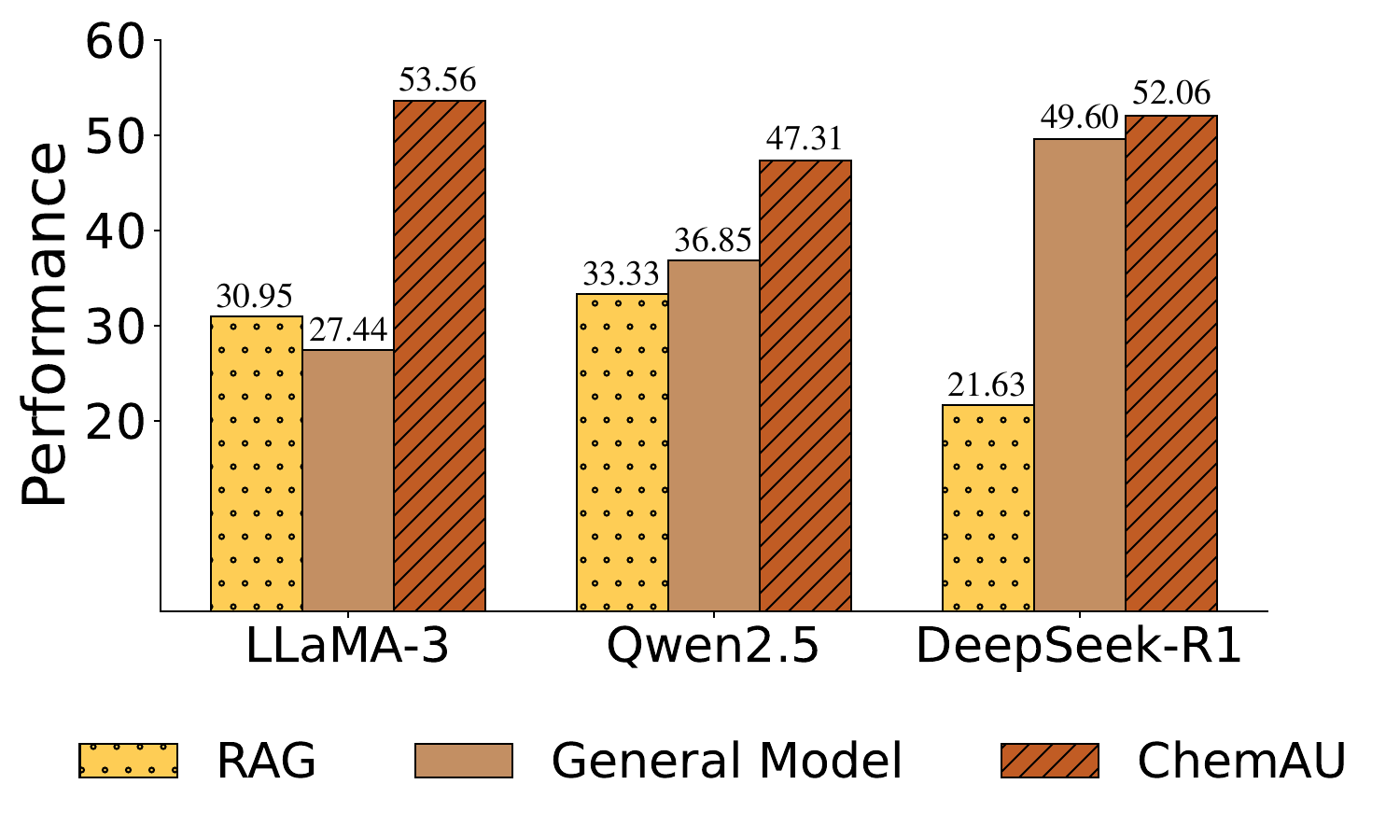}
        \caption{MMLU-Pro}
        \label{fig:reasoning_performance_comparison_mmlupro}
    \end{subfigure}
    \hfill
    \begin{subfigure}[b]{0.31\textwidth}
        \centering
        \includegraphics[width=\textwidth]{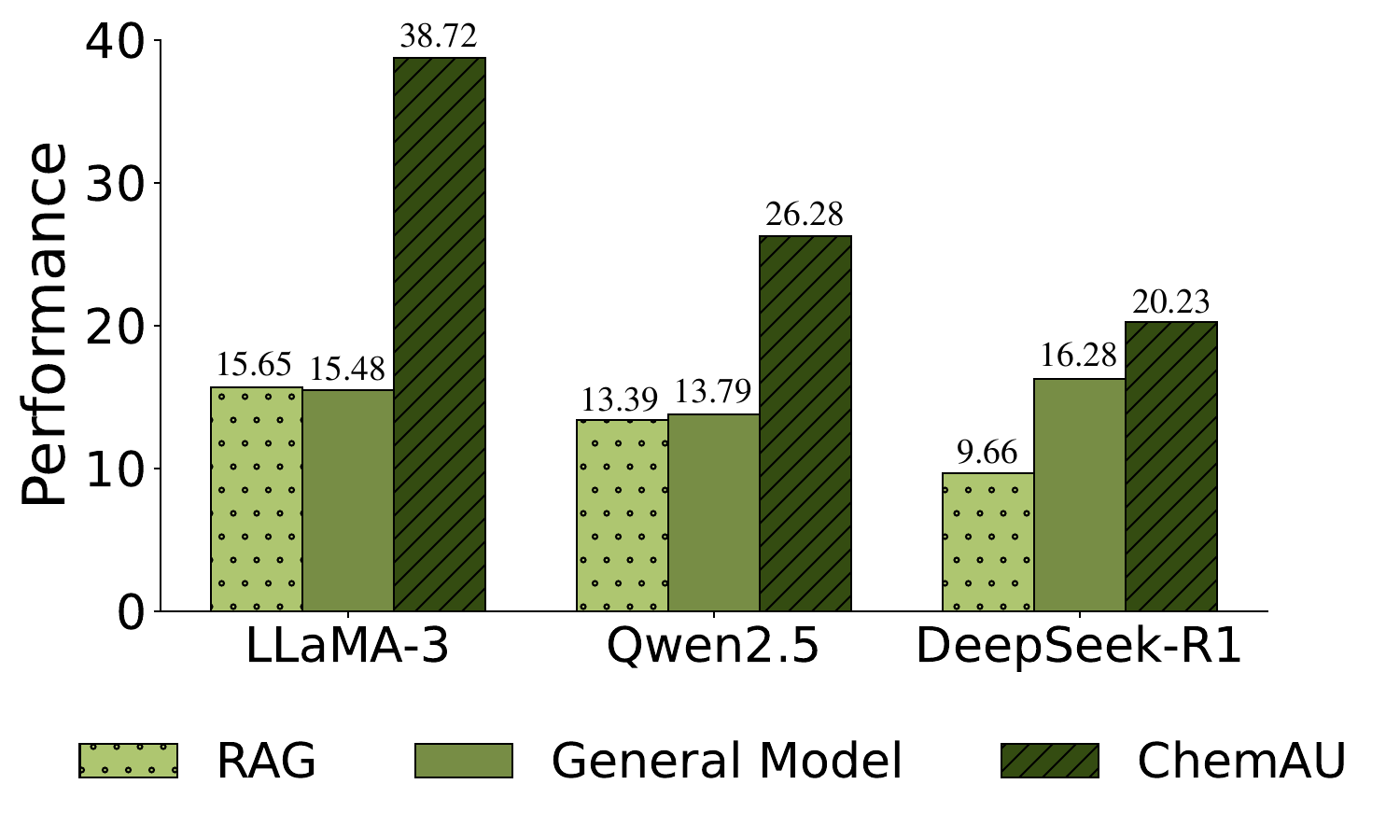}
        \caption{SuperGPQA}
        \label{fig:reasoning_performance_comparison_supergpqa}
    \end{subfigure}
    % \vspace{-0.2cm}
    \caption{Performance comparison of different reasoning approaches across three chemistry datasets. Each sub-figure shows the accuracy (\%) of three different reasoning methods (RAG, General LLMs only, and our proposed framework ChemAU) on a specific dataset. The results demonstrate that ChemAU consistently outperforms other approaches across all tested general models on all evaluation datasets. }
    \label{fig:reasoning_performance_comparison}
\end{figure}

\begin{itemize}
    \item \textbf{General LLMs Performance. }
In Table~\ref{tab:general_llms_performance}, we present the performance of general LLMs on chemistry reasoning tasks without any specialized enhancement. This baseline serves as a reference point to demonstrate the effectiveness of our uncertainty-driven adaptive reasoning framework. 

    \item \textbf{Domain-Specific Model Performance. }
We evaluate the performance of our constructed chemistry domain model by directly answering questions without the reasoning support of general LLMs. As shown in Table~\ref{tab:general_llms_performance}, when the domain model independently addresses complex chemistry problems, its accuracy rate is relatively low, strongly proving that the domain model's value lies not in directly enhancing the general model's reasoning process, but rather in providing specialized chemistry domain knowledge. 

    \item \textbf{Retrieval-Augmented Generation Approach. }
We explore the Retrieval-Augmented Generation (RAG) approach by retrieving relevant chemistry knowledge based on specific problems and feeding both the problems and retrieved knowledge to the general model to generate reasoning steps and answers. However, as shown in Figure~\ref{fig:reasoning_performance_comparison}, the performance using RAG is actually inferior to using the general model alone. This may primarily because the knowledge retrieved by RAG is too broad and may not precisely identify the domain knowledge that the model actually lacks or needs. When the provided knowledge is not closely relevant, it may mislead the model's reasoning process, resulting in decreased accuracy. This experiment further demonstrates the importance of precisely identifying and supplementing the critical domain knowledge that the general model lacks during the reasoning process, which is precisely the role of the domain model triggered by uncertainty detection. 
\end{itemize}

\begin{table}[htbp]
  \centering
  \caption{Performance Comparison of General LLMs and Domain-Specific Model on Chemistry Reasoning Tasks. This table presents the accuracy (\%) of three instruction-tuned open-source LLMs (\textit{LLaMA-3}, \textit{Qwen2.5}, \textit{DeepSeek-R1}) and the self-constructed chemistry domain model across three chemistry reasoning datasets. These results establish baseline performance for evaluating the effectiveness of our proposed adaptive uncertainty-driven reasoning framework. }
  \label{tab:general_llms_performance}

  \vspace{10pt}
  
  \begin{tabular}{ccccc} 
    \toprule
    Datasets & LLaMA-3 & Qwen2.5 & DeepSeek-R1 & Domain Model \\
    \midrule 
    GPQA & 20.43\% & 19.35\% & 22.58\% & 6.45\% \\
    MMLU-Pro & 27.44\% & 36.85\% & 49.60\% & 11.25\% \\
    SuperGPQA & 15.48\% & 13.79\% & 16.28\% & 9.89\% \\
    \bottomrule 
  \end{tabular}
  
\end{table}

\subsection{Performance Evaluation on Different LLM Backbones}
As illustrated in Figure~\ref{fig:reasoning_performance_comparison}, our method significantly improves the accuracy of general LLMs on chemistry domain problems. For instance, on the MMLU-Pro dataset, our framework achieves an accuracy of 53.56\% when using LLaMA-3 as general model, representing a 26.12\% improvement over using the model alone and 22.60\% improvement over using the RAG approach. Notably, this performance even surpasses that of the larger 14B parameter model, demonstrating the effectiveness of our framework. 

\subsection{Comparison of Uncertainty Estimation Methods}
\begin{figure}[htbp]
    \centering
    \begin{subfigure}[b]{0.31\textwidth}
        \centering
        \includegraphics[width=\textwidth]{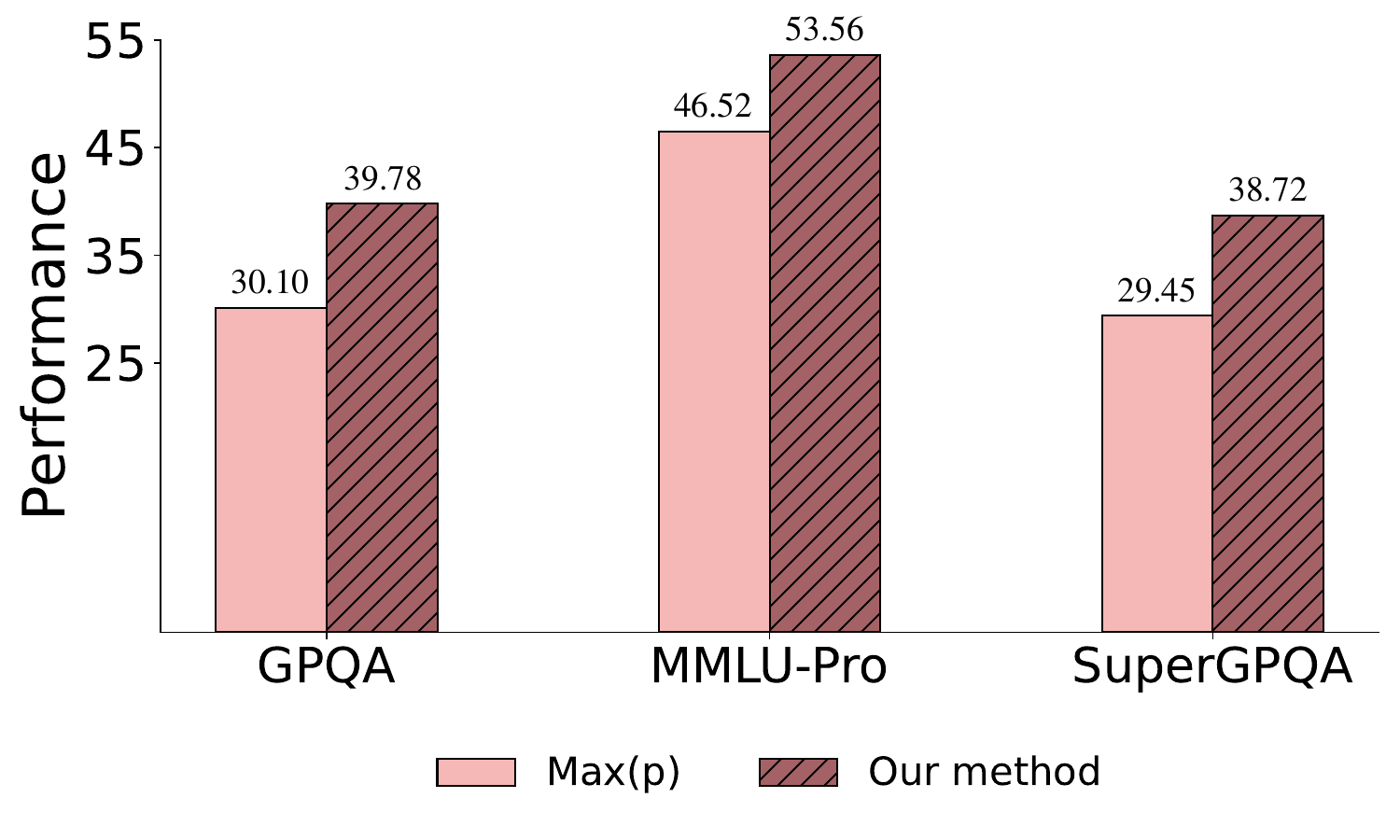}
        \caption{LLaMA-3}
        \label{fig:uncertainty_method_comparison_llama}
    \end{subfigure}
    \hfill
    \begin{subfigure}[b]{0.31\textwidth}
        \centering
        \includegraphics[width=\textwidth]{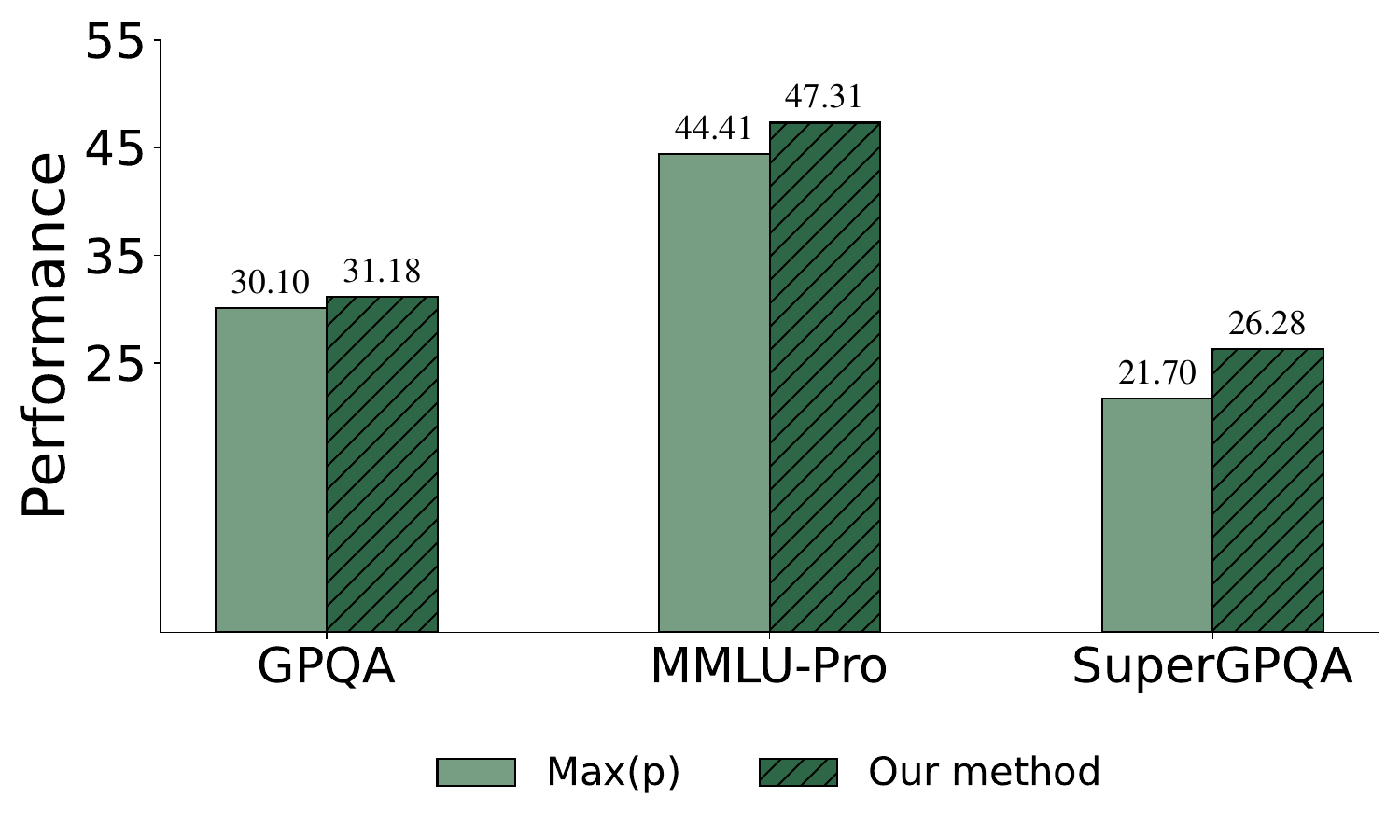}
        \caption{Qwen2.5}
        \label{fig:uncertainty_method_comparison_qwen}
    \end{subfigure}
    \hfill
    \begin{subfigure}[b]{0.31\textwidth}
        \centering
        \includegraphics[width=\textwidth]{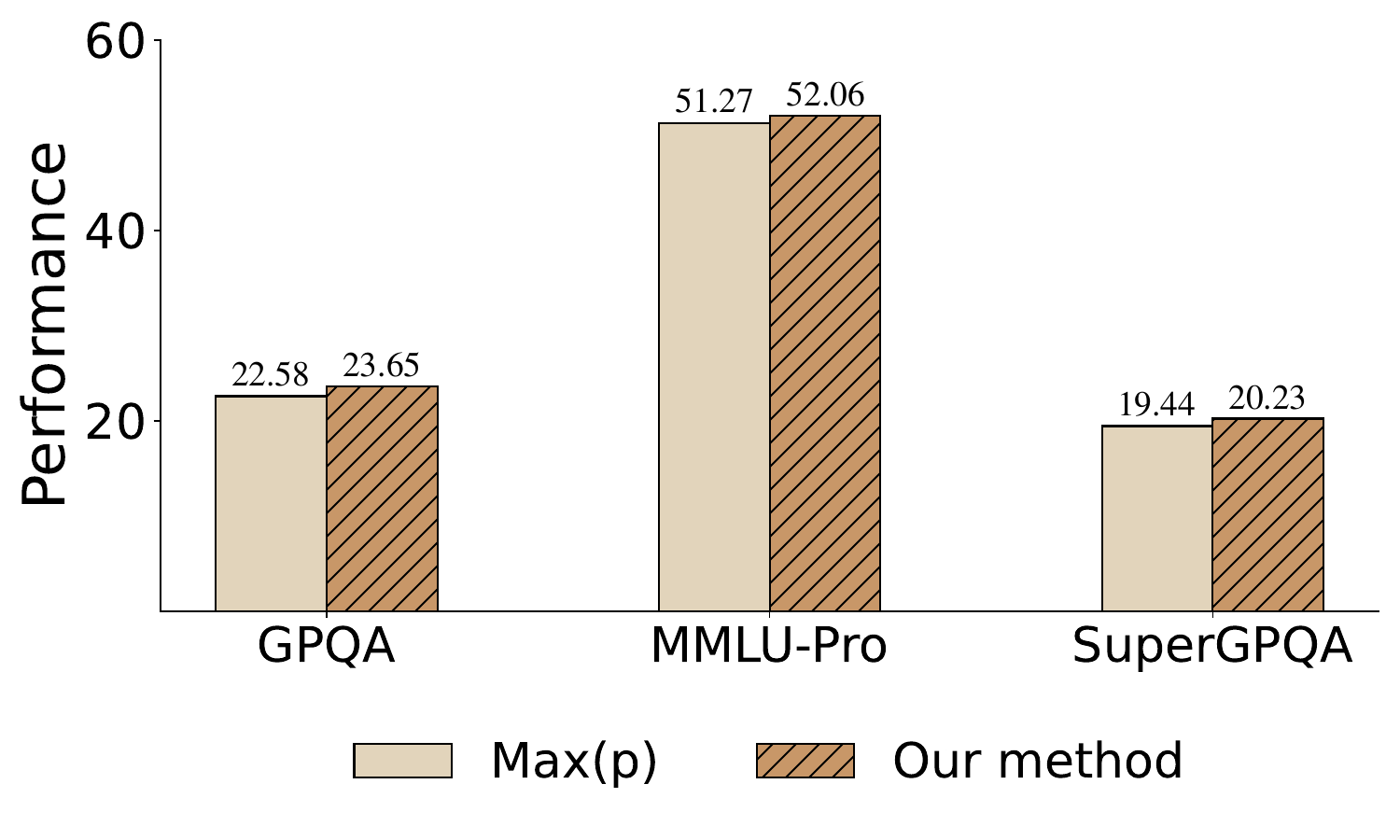}
        \vspace{-0.6cm}
        \caption{DeepSeek-R1}
        \label{fig:uncertainty_method_comparison_ds}
    \end{subfigure}
    \caption{Performance comparison of uncertainty estimation methods on chemistry reasoning tasks. The figure presents accuracy results (\%) across three general LLMs evaluated on three chemistry datasets. Our proposed dynamic uncertainty estimation method (dark bars) consistently outperforms the $Max(-logp)$ approach (light bars) across all experimental settings, with a maximum improvement of 9.68\%. The dynamic method demonstrates superior capability in identifying high-uncertainty reasoning steps where domain knowledge integration is required. }
    \label{fig:uncertainty_method_comparison}
\end{figure}

Figure~\ref{fig:uncertainty_method_comparison} illustrates the performance comparison between two uncertainty estimation methods within our framework. The results clearly demonstrate that our proposed dynamic uncertainty estimation method consistently outperforms $Max(-logp)$ across various general LLMs and datasets, substantially enhancing the overall framework performance. These findings confirm that our uncertainty estimation method can more precisely and promptly identify the missing domain knowledge in the reasoning chain, thereby effectively improving the model's accuracy. 

\section{Ablation Studies}

\label{sec:5}

\textbf{Necessity of the Domain Model. }
To investigate whether the specialized chemistry domain model provides chemistry knowledge that effectively compensates for the knowledge gaps in the general LLM, thereby enhancing its ability to solve chemistry reasoning problems, rather than the general model simply improving through self-reflection and iterative optimization to increase chemistry reasoning accuracy, we design and conduct an ablation experiment targeting the domain model component. This experiment aims to verify whether the key role of domain models lies in providing specialized chemistry knowledge that general models lack, rather than simply facilitating the general model's self-improvement process.

\begin{figure}[htbp]
    \centering
    \begin{subfigure}[b]{0.45\textwidth}
        \centering
        \includegraphics[width=\textwidth]{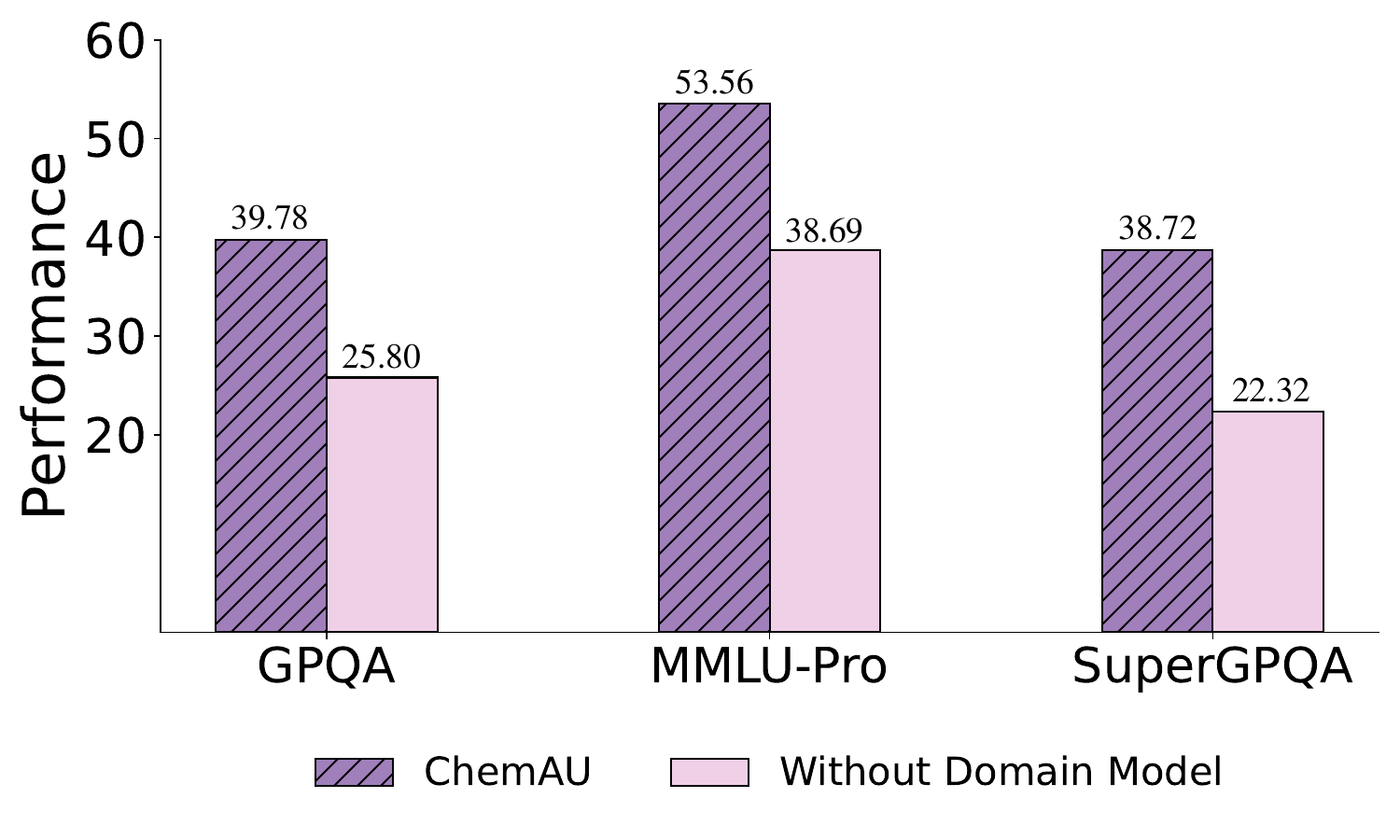}
        \label{fig:ablation_domain_model_llama}
    \end{subfigure}
    \hfill
    \begin{subfigure}[b]{0.45\textwidth}
        \centering
        \includegraphics[width=\textwidth]{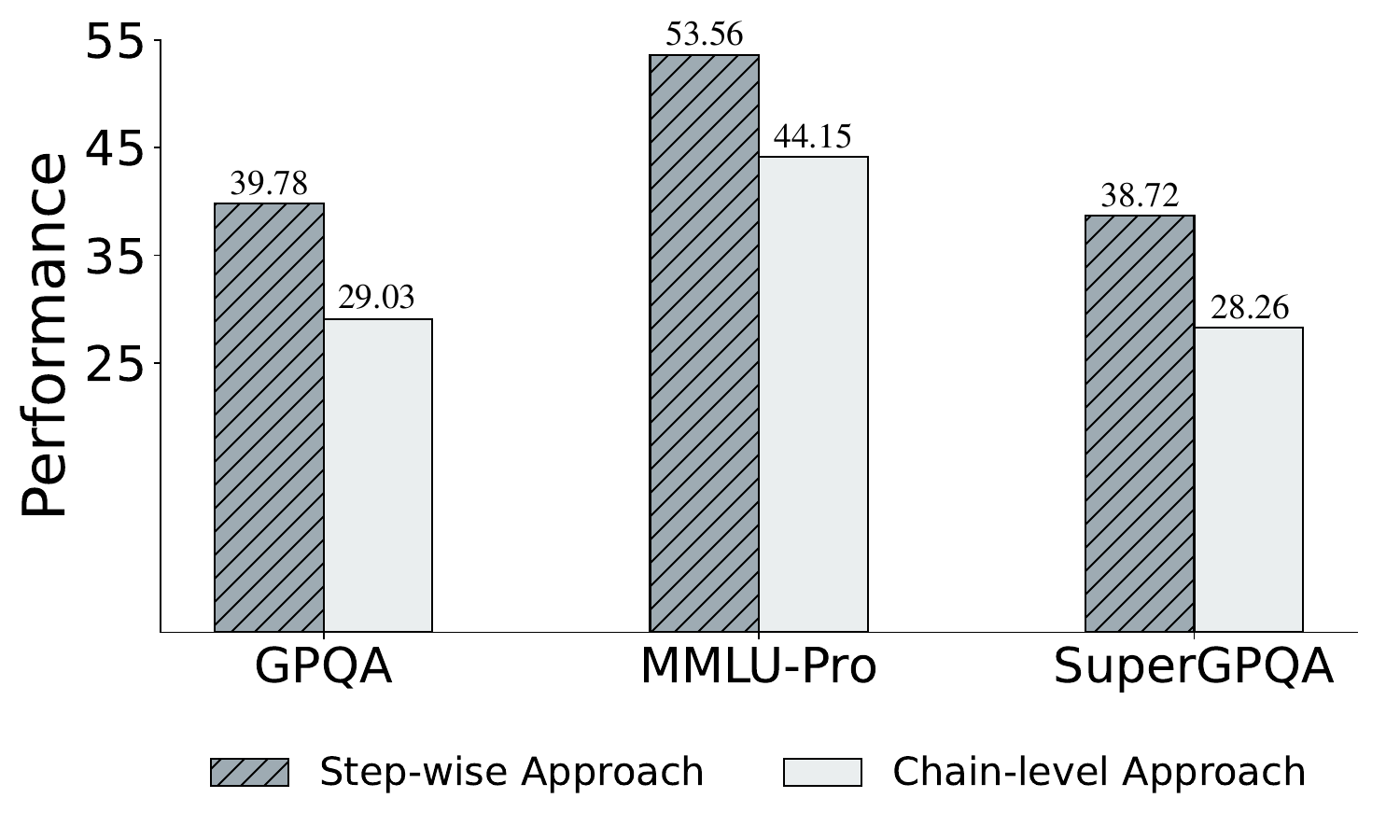}
        \label{fig:ablation_step_llama}
    \end{subfigure}
    \caption{(Left) The comparisons of the accuracy (\%) of our complete reasoning framework against a variant without the specialized chemistry domain model across different chemistry datasets. 
    The variant still identifies high-uncertainty reasoning steps but is resolved through iterative general model re-reasoning rather than domain knowledge integration. Results show significant performance degradation (up to 14.87\% accuracy drop) when domain expertise is removed. 
    (Right) The comparisons of the accuracy results (\%) of our proposed step-wise approach against a variant that treats the entire reasoning chain as a single unit across different general models and chemistry datasets. 
    In this chain-level approach, the complete reasoning chain is evaluated as one unit before domain knowledge integration, rather than assessing uncertainty at individual reasoning steps. Results demonstrate that the chain-level method diminishes the step-wise approach's effectiveness by an average of 10.20\%. }
    \label{fig:ablation_comparison}
\end{figure}

In this experiment, we remove the chemistry domain model from our proposed reasoning framework while maintaining the uncertainty estimation method to identify high-risk reasoning steps. 
Instead of leveraging domain expertise for knowledge correction, we provide the general model with only the original reasoning steps, potential error information, and the original chemistry problem, requesting it to regenerate the reasoning chain iteratively. 
As shown in Figure~\ref{fig:ablation_comparison}, this ablation study demonstrates a significant performance decline compared to our complete reasoning framework, with accuracy rates dropping by up to 16.39\%.
These results empirically validate that general LLMs, even when prompted with uncertainty information, struggle to independently correct domain-specific reasoning errors. 
They often possess knowledge gaps or conceptual misunderstandings in specialized domains that cannot be resolved through simple re-reasoning processes. 
Our findings strongly highlight the necessity of integrating domain-specific knowledge through specialized models, confirming that our proposed framework effectively bridges the gap between general reasoning capabilities and specialized domain expertise in solving complex chemistry problems.

\textbf{Necessity of Step-wise Uncertainty Detection. }
To validate the efficacy of our step-wise approach, we design an ablation experiment that examines the necessity of fine-grained uncertainty estimation and targeted knowledge injection during the reasoning process. Our proposed framework fundamentally innovates through its step-wise methodology that both identifies specific reasoning steps with high uncertainty and precisely supplements chemistry-specific knowledge exactly where needed. We hypothesize that treating the entire reasoning chain as a unit may compromise both uncertainty detection and knowledge supplementation processes. When evaluated as a whole, critical knowledge deficiencies at individual steps might be obscured, and the domain model may not be able to provide precise, step-specific knowledge corrections. This could lead to either overlooking crucial knowledge gaps or introducing broadly generalized information that fails to address specific reasoning errors, ultimately resulting in less effective knowledge integration and compromising the overall reasoning quality. 

In this ablation study, instead of performing uncertainty estimation on individual reasoning steps, we treat the entire reasoning chain as a single unit, which is directly fed into the domain model to obtain relevant knowledge. 
Subsequently, the original question, along with this knowledge supplement, is reintroduced to the general model to regenerate the entire reasoning chain. As illustrated in Figure~\ref{fig:ablation_comparison}, this variant performs worse than our proposed step-wise framework, which validates that fine-grained uncertainty detection at each reasoning step enables more precise identification of knowledge gaps in the general model's reasoning. 
The step-wise approach allows for targeted and timely domain knowledge injection exactly where it is needed, avoiding overwhelming the model with potentially irrelevant domain information. 

\section{Conclusion}

\label{sec:6}

In this paper, we observe that LLMs often suffer from hallucinations and poor performance when answering domain-specific questions due to insufficient specialized knowledge. To address this issue, we propose a reasoning framework for the chemistry domain that incorporates a novel adaptive uncertainty estimation method. Experimental results demonstrate that our framework significantly improves the accuracy of LLMs on chemistry-related questions, enabling smaller models to achieve or even exceed the performance of models with larger parameter configurations.
\textbf{Limitations and Future Work. }
The proposed adaptive uncertainty estimation method is only applicable to open-source LLMs. For black-box models, this method cannot be directly applied since it's impossible to access the logit values of generated tokens. Future research could develop more universally methods.

\newpage
\section*{Appendix}
\appendix

\section{Performance between Different Uncertainty Estimation Methods}

In this section, we compare existing token-based uncertainty estimation methods for open-source LLMs. Token-based uncertainty estimation methods typically utilize the logit values of tokens provided by open-source LLMs. Following statistical principles, the basic method is to multiply the logit values of all generated tokens to reflect the overall confidence of the LLM in the entire generated sentence. The formula can be expressed as follows: 
\begin{equation}
    Base(i) = \prod_{j=1}^{L_i}p_{ij}, 
\end{equation}
where $L_i$ represents the length of generated tokens, and $p_{ij}$ represents the logit value of each token. For consistent expression, we convert it to logarithmic form, which can be expressed as follows:
\begin{equation}
    Log_{Base}(i) = \sum_{j=1}^{L_i} log(p_{ij}).
\end{equation}

\begin{figure}[t]
    \centering
    \includegraphics[width=\textwidth]{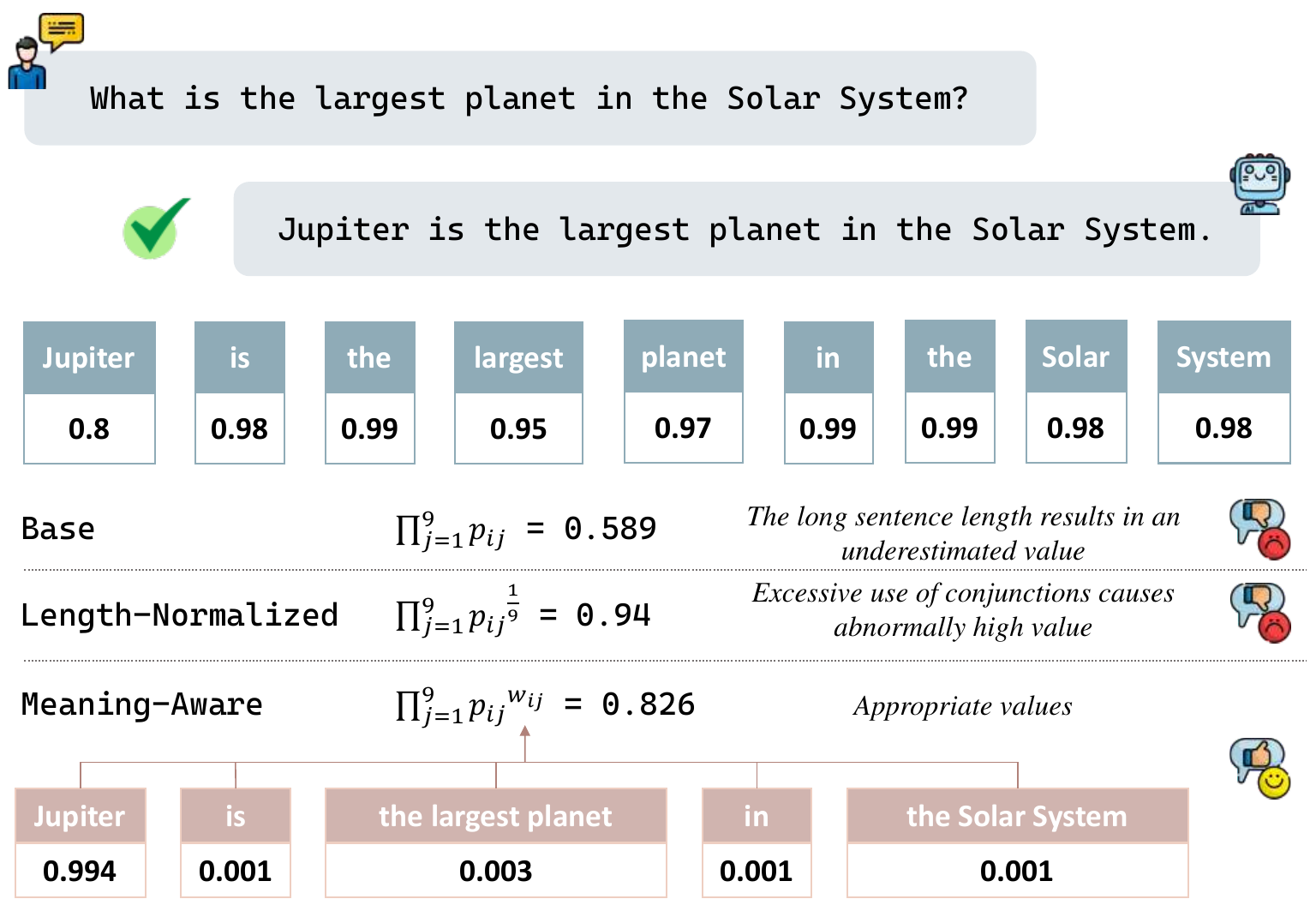}
    \caption{Different uncertainty estimation methods demonstrate varying performance on the  same sentence in everyday language corpora. The Base method is sensitive to sentence length influence, often resulting in underestimated uncertainty values that may incorrectly classify reliable answers as unreliable. While the Length-Normalized method averages values based on token count from the Base method, the abundance of conjunctions with high logit values tends to produce overestimated uncertainty values, potentially misclassifying unreliable answers as trustworthy. In contrast, the Meaning-Aware method calculates weights according to each token's semantic contribution, yielding more reasonable and accurate uncertainty assessment values. }
    \label{fig:Appendix_Uncertainty_comparison}
\end{figure}

However, it typically exhibits poor performance when evaluating long-form responses, as the product of token probabilities inherently diminishes with response length, even when these longer responses are semantically equivalent to their shorter counterparts. To mitigate this limitation, researchers have developed several variant methods to effectively reduce the dependency between the metrics and the sequence length. The comparison of these methods is shown in Figure~\ref{fig:Appendix_Uncertainty_comparison}. 

\subsection{Length-Normalized Scoring Function}
The length-normalized method aims to distribute uncertainty across each token to mitigate the impact of sentence length on uncertainty calculations~\cite{manakul2023selfcheckgpt}. This method can be mathematically expressed as follows:
\begin{equation}
    LN(i) = \prod_{j=1}^{L_i}{p_{ij}}^{\frac{1}{L_i}}. 
\end{equation}

The corresponding logarithmic form is: 
\begin{equation}
    Log_{LN}(i) = \frac{1}{L_i} \sum_{j=1}^{L_i} log(p_{ij}), 
\end{equation}

which is essentially identical to the base method mentioned above, with the only difference being that it applies an averaging process according to sentence length, ensuring that sentences of varying lengths can be evaluated using comparable uncertainty assessment criteria.

\subsection{Semantic Contribution Weighting}

The length-normalized approach treats each token equally, meaning they contribute identically to the uncertainty estimation value. In reality, different words can have varying impacts on a sentence's meaning within the question context, particularly for connective words such as ``the,'' ``an,'' ``of,'' and similar tokens, while the key tokens are the ones truly answering the question. Therefore, some researchers suggest that when calculating uncertainty values, tokens should be assigned different weights based on their semantic contribution level. They typically employ additional semantic detection models to specifically compare the semantic similarity between the original sentence and the sentence with certain tokens removed. If the similarity is high, it indicates that the token does not significantly affect the sentence's meaning, thus, it should be assigned a lower weight in uncertainty estimation. If the similarity is low, it suggests that the token is crucial to the essential meaning of the sentence, therefore, it should be given a higher weight in uncertainty estimation~\cite{duan2023shifting}. The specific formula is as follows:
\begin{equation}
    SCW(i) = \prod_{j=1}^{L_i}{p_{ij}}^{w_{ij}}, 
\end{equation}

where $w_{ij}$ represents the weight of token $p_{ij}$. The corresponding logarithmic form is: 
\begin{equation}
    Log_{SCW}(i) = w_{ij} \sum_{j=1}^{L_i} log(p_{ij}). 
\end{equation}

\begin{figure}[t]
    \centering
    \includegraphics[width=\textwidth]{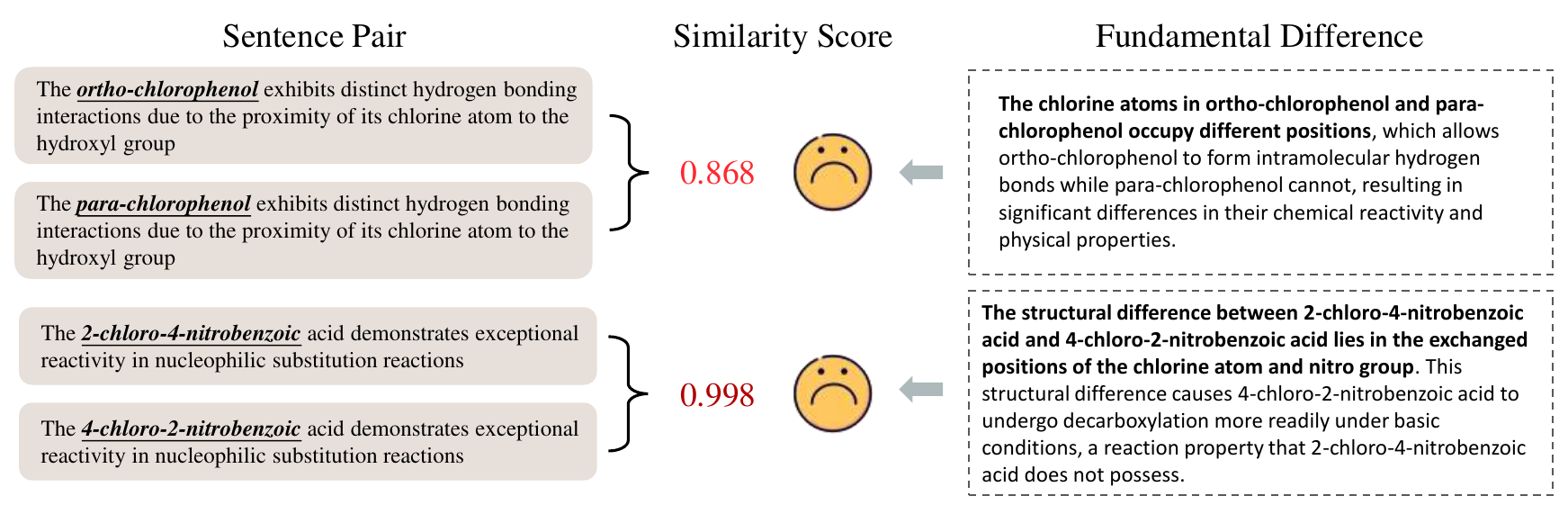}
    \caption{Models designed to assess semantic similarity perform poorly when evaluating sentences containing specific chemical terminology. We utilize the \textbf{\textit{sentence-transformers/all-MiniLM-L6-v2}} for testing. The two examples above clearly illustrate this issue: despite these sentences having fundamental differences in chemical expressions that should classify them as completely unrelated, the model fails to identify their significant distinctions and instead incorrectly evaluate them as semantically almost identical. This indicates notable limitations in current models when processing text from specialized chemistry domains. }
    \label{fig:Appendix_Semantic}
\end{figure}

However, this method performs poorly on chemistry problems for the same reason we mentioned in the main text regarding why LLMs don't perform as well on chemistry questions as they do on general topics: semantic similarity detection models are primarily trained on everyday general corpora, with relatively limited chemistry-related content. Consequently, these models cannot accurately identify the semantic impact and importance of key chemistry tokens within a sentence. More critically, even when key chemistry tokens express entirely different meanings, these models fail to detect the semantic differences, as illustrated in Figure~\ref{fig:Appendix_Semantic}. 

\section{Computer Resources and Experiment Details}

For both general LLMs and the chemistry domain model, inference is performed with a temperature of 0.3 and top-k sampling, retrieving 4 candidate tokens per position.
The general LLM is configured with a maximum sequence length of 1024 tokens, while the chemistry domain model uses a reduced maximum sequence length of 100 tokens. GPU memory utilization is set to 0.6 for the general LLM and 0.2 for the chemistry domain model. 
The basic threshold value is set to -1.5, while the hyperparameter $\alpha$, which incorporates the relative positions of reasoning steps into the uncertainty estimation model, is set to -0.08. All experiments are conducted using NVIDIA A100 (80GB) GPUs.

\section{Prompt Template}

\begin{figure}[t]
    \centering
    \includegraphics[width=\textwidth]{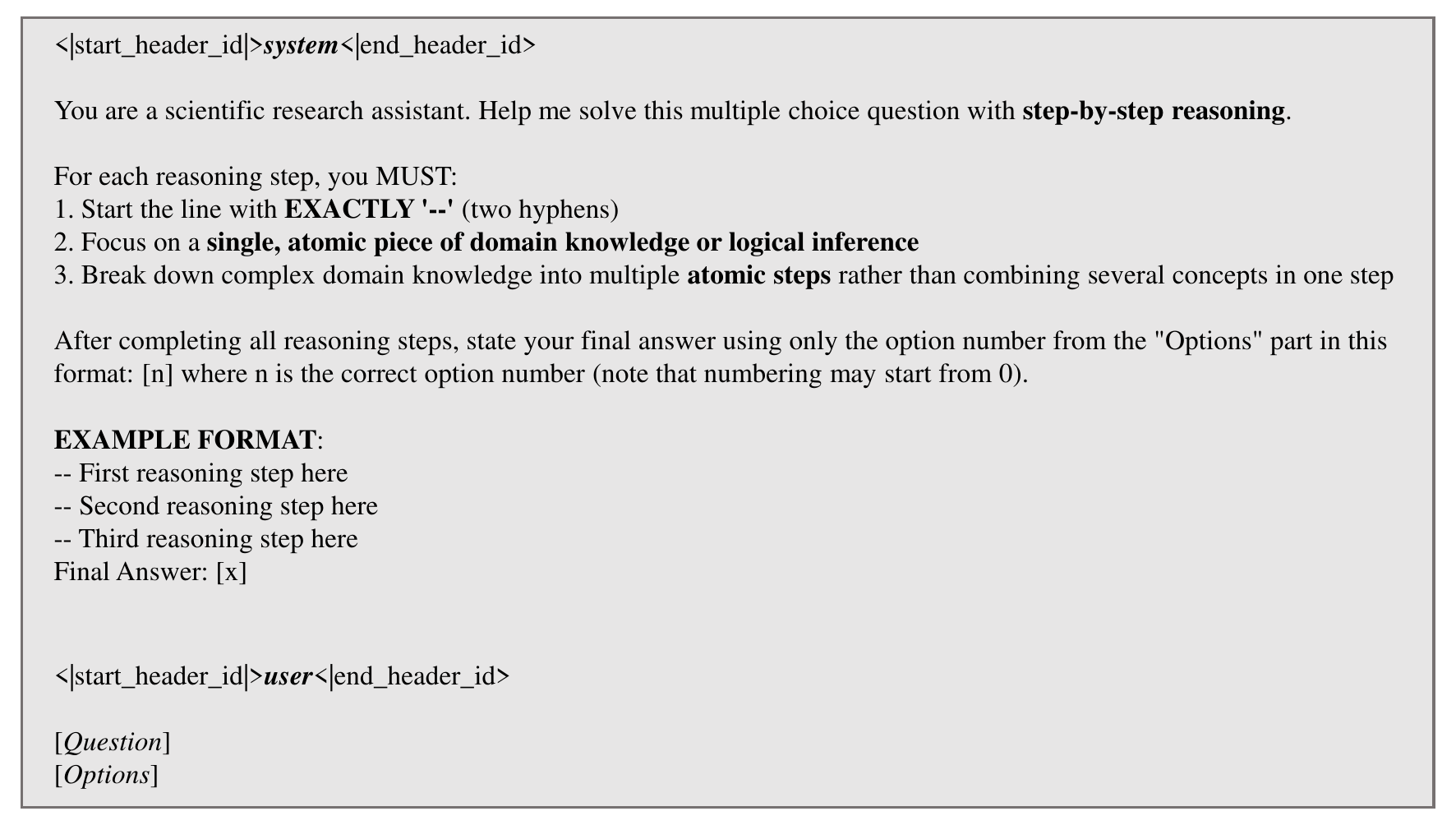}
    \caption{A prompt template designed to guide the general LLM in generating an initial reasoning chain for chemistry problems. In this template, the "step-by-step reasoning" instruction encourages the model to generate CoT style reasoning processes, using double hyphens (`-\ -') as the starting identifier for each reasoning step, and emphasizing that each reasoning step should focus on atomic chemistry knowledge points. This fine-grained division facilitates the subsequent uncertainty estimation method in identifying and locating the missing chemistry knowledge in general LLMs. ``Question'' and ``Options'' part in user should be filled with the chemistry problem. \textit{Since the DeepSeek-R1 series models do not recommend the separation of system and user roles, the system content is directly integrated into the user section}. }
    \label{fig:Template General Begin}
\end{figure}

\begin{figure}[t]
    \centering
    \includegraphics[width=\textwidth]{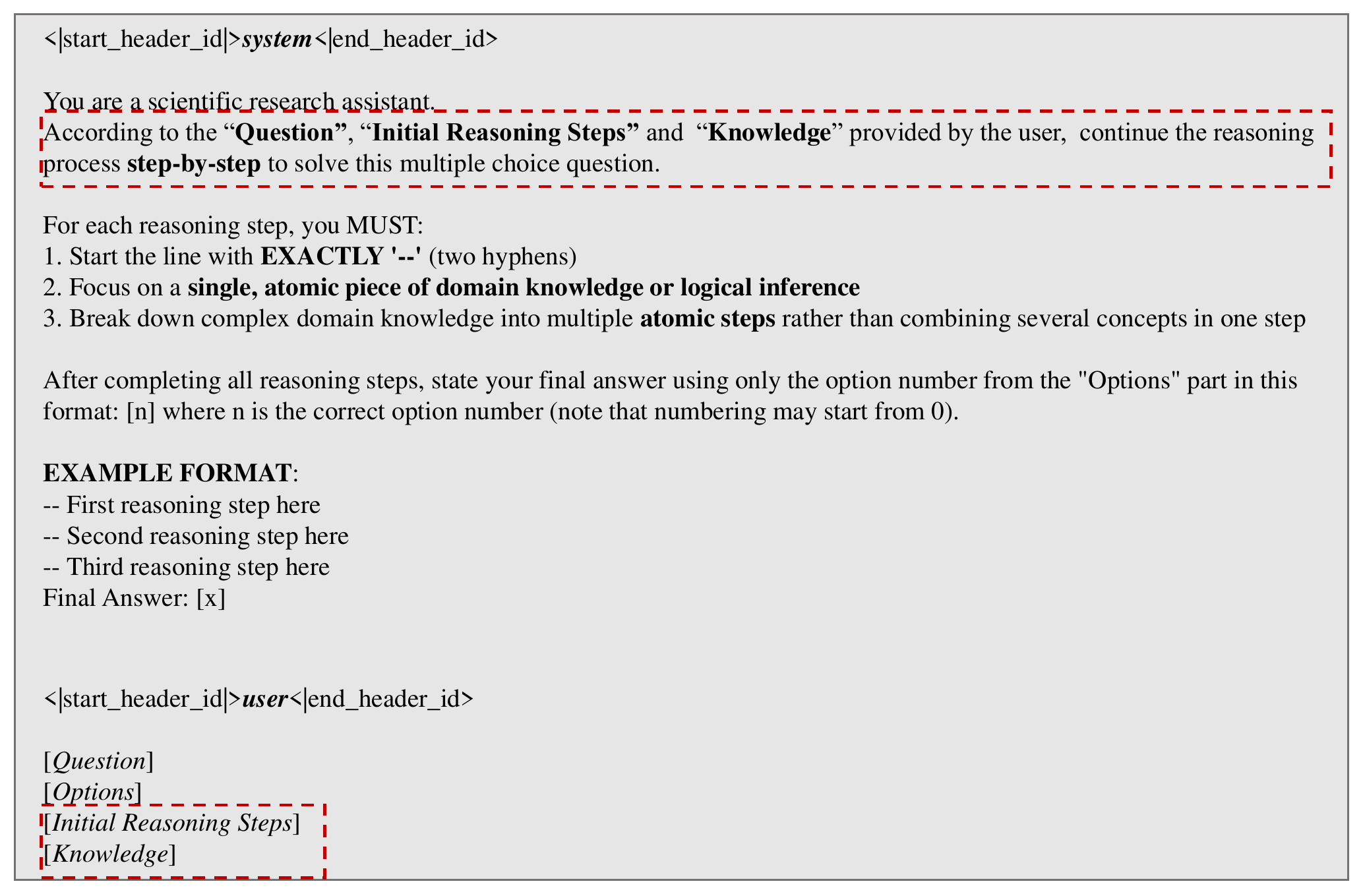}
    \caption{A prompt template designed to guide the general LLM in regenerating the reasoning chain for chemistry problems according to the ``\textit{Initial Reasoning Steps}'' and ``\textit{Knowledge}''. This template is based on the previous template with modifications only to the sections outlined in red. \textit{Since the DeepSeek-R1 series models do not recommend the separation of system and user roles, the system content is directly integrated into the user section}. }
    \label{fig:Template General Regenerate}
\end{figure}

For all experiments, we employ identical prompt templates to ensure fairness and reliability in our evaluations. 
Specifically, we utilize two distinct prompt templates throughout our experiments. The first is for guiding general LLMs to generate reasoning on initial chemistry problems. The second is for reintroducing the acquired knowledge, correct reasoning steps, and original chemistry questions back into the general LLM to continue generating more accurate reasoning processes, as shown in Figures~\ref{fig:Template General Begin} and~\ref{fig:Template General Regenerate}. 

\section{Addition to Experiment and Ablation Section}

\begin{figure}[H]
    \centering
    \includegraphics[width=\textwidth]{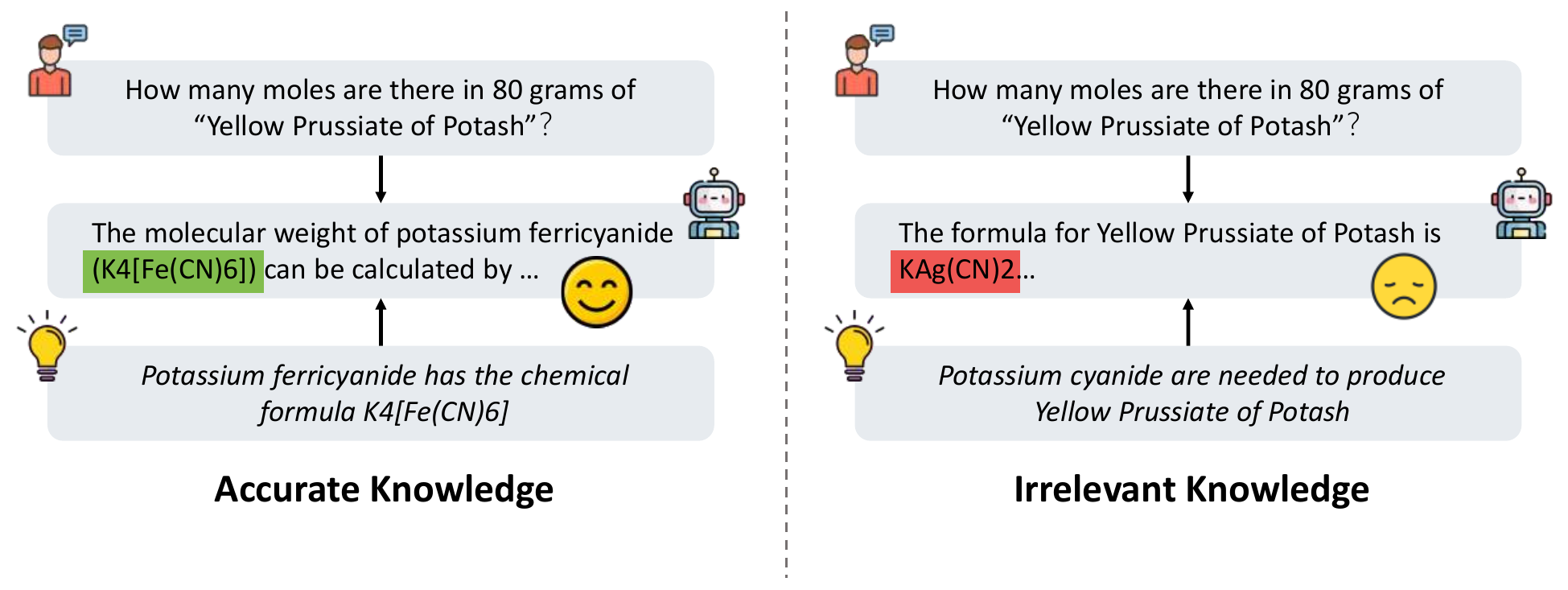}
    \caption{Knowledge Augmentation Comparison. Precise and relevant knowledge augmentation facilitates correct reasoning in models, while broad or irrelevant knowledge augmentation tends to induce model hallucinations, consequently leading to erroneous reasoning. }
    \label{fig:RAG Comparison}
\end{figure}

Figure~\ref{fig:RAG Comparison} demonstrates the comparison of different knowledge augmentation strategies. 
Precise and relevant knowledge can effectively guide models toward correct reasoning, while broad or irrelevant knowledge misleads models, resulting in erroneous reasoning outcomes. 
This finding explains why the RAG approach performed worse than using the general LLM alone in our experiments. 

\begin{figure}[t]
    \centering
    \begin{subfigure}[b]{0.45\textwidth}
        \centering
        \includegraphics[width=\textwidth]{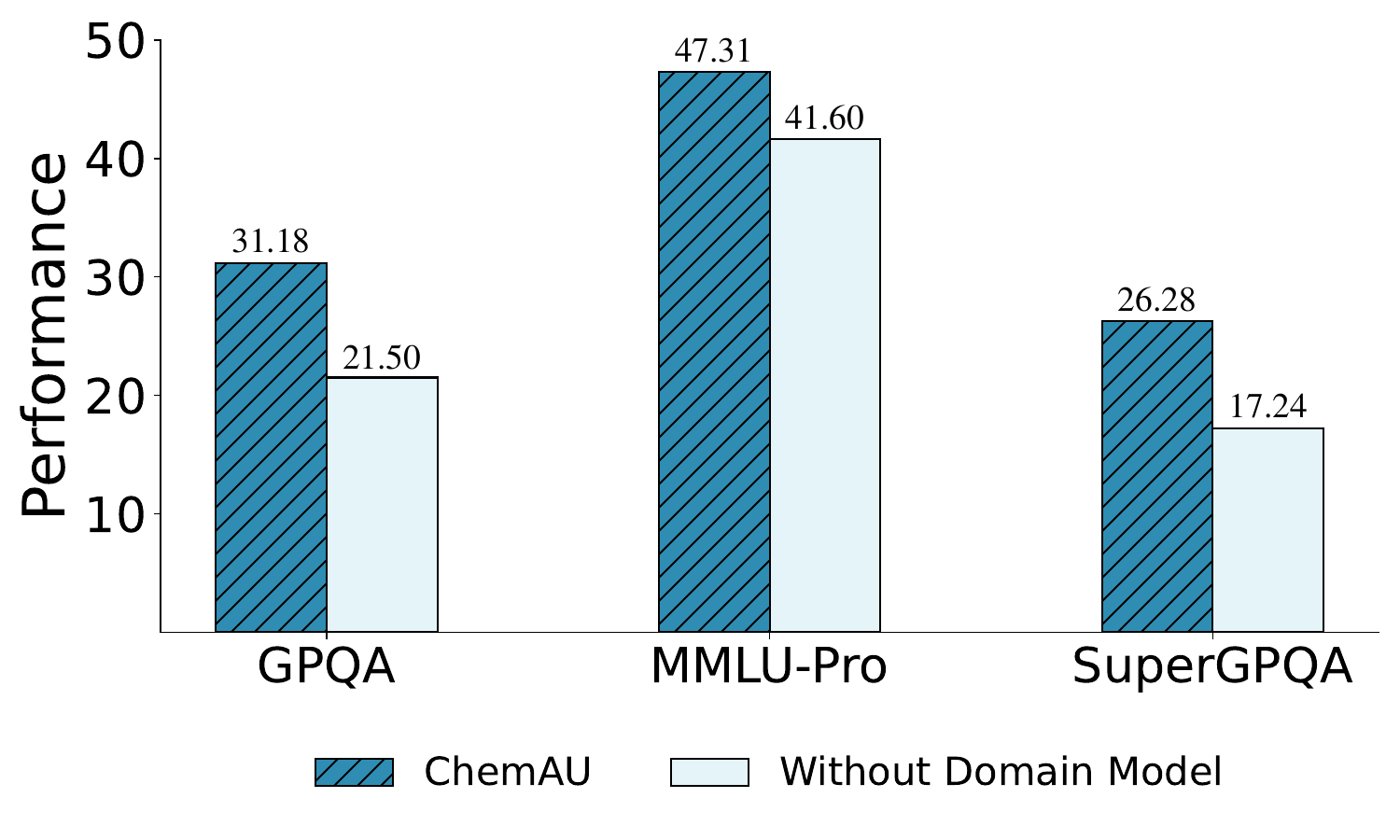}
        \label{fig:ablation_domain_model_qwen}
    \end{subfigure}
    \hfill
    \begin{subfigure}[b]{0.45\textwidth}
        \centering
        \includegraphics[width=\textwidth]{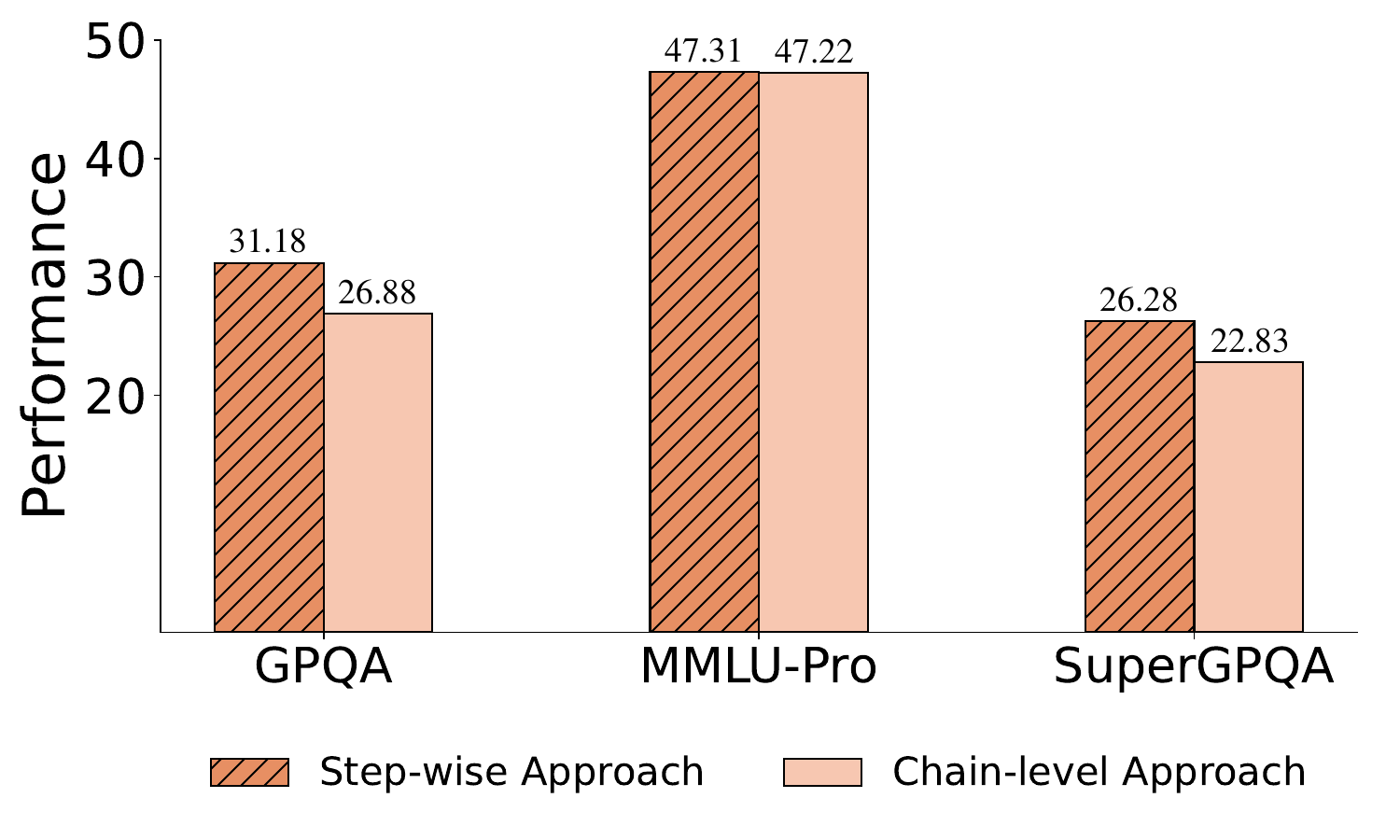}
        \label{fig:ablation_step_qwen}
    \end{subfigure}
    \caption{(Left) The comparisons of the accuracy (\%) of our complete reasoning framework against a variant without the specialized chemistry domain model across different chemistry datasets with \textit{Qwen2.5} as general LLM. 
    (Right) The comparisons of the accuracy results (\%) of our proposed step-wise approach against a variant that treats the entire reasoning chain as a single unit across different general models and chemistry datasets with \textit{Qwen2.5} as general LLM. }
    \label{fig:ablation_comparison_qwen}
\end{figure}

For the ablation experiments, we still conduct ablation studies on the domain model and step-wise uncertainty detection using \textit{Qwen2.5} as the general LLM, with results shown in Figure~\ref{fig:ablation_comparison_qwen}.
Consistent with the results obtained using LLaMA-3 as the general model in the main text, both ablation experiments yield lower performance than the complete framework, which aligns with the conclusions drawn in Section 5 in the main text. 

\FloatBarrier

\newpage
\bibliographystyle{unsrtnat}
\bibliography{references}  %%% Uncomment this line and comment out the ``thebibliography'' section below to use the external .bib file (using bibtex) .

\begin{thebibliography}{52}
\providecommand{\natexlab}[1]{#1}
\providecommand{\url}[1]{\texttt{#1}}
\expandafter\ifx\csname urlstyle\endcsname\relax
  \providecommand{\doi}[1]{doi: #1}\else
  \providecommand{\doi}{doi: \begingroup \urlstyle{rm}\Url}\fi

\bibitem[Topsakal and Akinci(2023)]{topsakal2023creating}
Oguzhan Topsakal and Tahir~Cetin Akinci.
\newblock Creating large language model applications utilizing langchain: A primer on developing llm apps fast.
\newblock In \emph{International Conference on Applied Engineering and Natural Sciences}, volume~1, pages 1050--1056, 2023.

\bibitem[Achiam et~al.(2023)Achiam, Adler, Agarwal, Ahmad, Akkaya, Aleman, Almeida, Altenschmidt, Altman, Anadkat, et~al.]{achiam2023gpt}
Josh Achiam, Steven Adler, Sandhini Agarwal, Lama Ahmad, Ilge Akkaya, Florencia~Leoni Aleman, Diogo Almeida, Janko Altenschmidt, Sam Altman, Shyamal Anadkat, et~al.
\newblock Gpt-4 technical report.
\newblock \emph{arXiv preprint arXiv:2303.08774}, 2023.

\bibitem[Guo et~al.(2025)Guo, Yang, Zhang, Song, Zhang, Xu, Zhu, Ma, Wang, Bi, et~al.]{guo2025deepseek}
Daya Guo, Dejian Yang, Haowei Zhang, Junxiao Song, Ruoyu Zhang, Runxin Xu, Qihao Zhu, Shirong Ma, Peiyi Wang, Xiao Bi, et~al.
\newblock Deepseek-r1: Incentivizing reasoning capability in llms via reinforcement learning.
\newblock \emph{arXiv preprint arXiv:2501.12948}, 2025.

\bibitem[Zubiaga(2024)]{zubiaga2024natural}
Arkaitz Zubiaga.
\newblock Natural language processing in the era of large language models, 2024.

\bibitem[Sapkota et~al.(2024)Sapkota, Meng, and Karkee]{sapkota2024synthetic}
Ranjan Sapkota, Zhichao Meng, and Manoj Karkee.
\newblock Synthetic meets authentic: Leveraging llm generated datasets for yolo11 and yolov10-based apple detection through machine vision sensors.
\newblock \emph{Smart Agricultural Technology}, 9:\penalty0 100614, 2024.

\bibitem[Li et~al.(2024{\natexlab{a}})Li, Chen, Yang, Ai, Jia, Liu, Lin, Wu, Yuan, Hu, et~al.]{li2024legalagentbench}
Haitao Li, Junjie Chen, Jingli Yang, Qingyao Ai, Wei Jia, Youfeng Liu, Kai Lin, Yueyue Wu, Guozhi Yuan, Yiran Hu, et~al.
\newblock Legalagentbench: Evaluating llm agents in legal domain.
\newblock \emph{arXiv preprint arXiv:2412.17259}, 2024{\natexlab{a}}.

\bibitem[Goyal et~al.(2024)Goyal, Rastogi, Rajagopal, Yuan, Zhao, Chintagunta, Naik, and Ward]{goyal2024healai}
Sagar Goyal, Eti Rastogi, Sree~Prasanna Rajagopal, Dong Yuan, Fen Zhao, Jai Chintagunta, Gautam Naik, and Jeff Ward.
\newblock Healai: A healthcare llm for effective medical documentation.
\newblock In \emph{Proceedings of the 17th ACM International Conference on Web Search and Data Mining}, pages 1167--1168, 2024.

\bibitem[Wei et~al.(2022)Wei, Wang, Schuurmans, Bosma, Xia, Chi, Le, Zhou, et~al.]{wei2022chain}
Jason Wei, Xuezhi Wang, Dale Schuurmans, Maarten Bosma, Fei Xia, Ed~Chi, Quoc~V Le, Denny Zhou, et~al.
\newblock Chain-of-thought prompting elicits reasoning in large language models.
\newblock \emph{Advances in neural information processing systems}, 35:\penalty0 24824--24837, 2022.

\bibitem[Renze and Guven(2024)]{renze2024self}
Matthew Renze and Erhan Guven.
\newblock Self-reflection in llm agents: Effects on problem-solving performance.
\newblock \emph{arXiv preprint arXiv:2405.06682}, 2024.

\bibitem[Qian et~al.(2023)Qian, Tang, Yang, Liang, and Liu]{qian2023can}
Chen Qian, Huayi Tang, Zhirui Yang, Hong Liang, and Yong Liu.
\newblock Can large language models empower molecular property prediction?
\newblock \emph{arXiv preprint arXiv:2307.07443}, 2023.

\bibitem[Huang et~al.(2024{\natexlab{a}})Huang, Qu, Cousins, Johnson, Yin, Shah, Zhou, Altman, Wang, and Cong]{huang2024crispr}
Kaixuan Huang, Yuanhao Qu, Henry Cousins, William~A Johnson, Di~Yin, Mihir Shah, Denny Zhou, Russ Altman, Mengdi Wang, and Le~Cong.
\newblock Crispr-gpt: An llm agent for automated design of gene-editing experiments.
\newblock \emph{arXiv preprint arXiv:2404.18021}, 2024{\natexlab{a}}.

\bibitem[Wang et~al.(2019)Wang, Guo, Wang, Sun, and Huang]{wang2019smiles}
Sheng Wang, Yuzhi Guo, Yuhong Wang, Hongmao Sun, and Junzhou Huang.
\newblock Smiles-bert: large scale unsupervised pre-training for molecular property prediction.
\newblock In \emph{Proceedings of the 10th ACM international conference on bioinformatics, computational biology and health informatics}, pages 429--436, 2019.

\bibitem[Honda et~al.(2019)Honda, Shi, and Ueda]{honda2019smiles}
Shion Honda, Shoi Shi, and Hiroki~R Ueda.
\newblock Smiles transformer: Pre-trained molecular fingerprint for low data drug discovery.
\newblock \emph{arXiv preprint arXiv:1911.04738}, 2019.

\bibitem[Bagal et~al.(2021)Bagal, Aggarwal, Vinod, and Priyakumar]{bagal2021molgpt}
Viraj Bagal, Rishal Aggarwal, PK~Vinod, and U~Deva Priyakumar.
\newblock Molgpt: molecular generation using a transformer-decoder model.
\newblock \emph{Journal of chemical information and modeling}, 62\penalty0 (9):\penalty0 2064--2076, 2021.

\bibitem[Axelrod and Gomez-Bombarelli(2022)]{axelrod2022geom}
Simon Axelrod and Rafael Gomez-Bombarelli.
\newblock Geom, energy-annotated molecular conformations for property prediction and molecular generation.
\newblock \emph{Scientific Data}, 9\penalty0 (1):\penalty0 185, 2022.

\bibitem[Zhang et~al.(2025)Zhang, Ding, Lv, Wang, Yin, Zhang, Yu, Wang, Li, Xiang, et~al.]{zhang2025scientific}
Qiang Zhang, Keyan Ding, Tianwen Lv, Xinda Wang, Qingyu Yin, Yiwen Zhang, Jing Yu, Yuhao Wang, Xiaotong Li, Zhuoyi Xiang, et~al.
\newblock Scientific large language models: A survey on biological \& chemical domains.
\newblock \emph{ACM Computing Surveys}, 57\penalty0 (6):\penalty0 1--38, 2025.

\bibitem[Zhang et~al.(2024{\natexlab{a}})Zhang, Liu, Tan, Chen, Yan, Yan, Li, Huang, Yue, Ouyang, et~al.]{zhang2024chemllm}
Di~Zhang, Wei Liu, Qian Tan, Jingdan Chen, Hang Yan, Yuliang Yan, Jiatong Li, Weiran Huang, Xiangyu Yue, Wanli Ouyang, et~al.
\newblock Chemllm: A chemical large language model.
\newblock \emph{arXiv preprint arXiv:2402.06852}, 2024{\natexlab{a}}.

\bibitem[Li et~al.(2025)Li, Zhang, Wang, Hao, Lei, Tan, Zhou, Liu, Yang, Xiong, et~al.]{li2025chemvlm}
Junxian Li, Di~Zhang, Xunzhi Wang, Zeying Hao, Jingdi Lei, Qian Tan, Cai Zhou, Wei Liu, Yaotian Yang, Xinrui Xiong, et~al.
\newblock Chemvlm: Exploring the power of multimodal large language models in chemistry area.
\newblock In \emph{Proceedings of the AAAI Conference on Artificial Intelligence}, volume~39, pages 415--423, 2025.

\bibitem[Asai et~al.(2023)Asai, Wu, Wang, Sil, and Hajishirzi]{asai2023self}
Akari Asai, Zeqiu Wu, Yizhong Wang, Avirup Sil, and Hannaneh Hajishirzi.
\newblock Self-rag: Learning to retrieve, generate, and critique through self-reflection.
\newblock In \emph{The Twelfth International Conference on Learning Representations}, 2023.

\bibitem[Liu et~al.(2025)Liu, Chen, Da, Chen, Lin, and Wei]{liu2025uncertainty}
Xiaoou Liu, Tiejin Chen, Longchao Da, Chacha Chen, Zhen Lin, and Hua Wei.
\newblock Uncertainty quantification and confidence calibration in large language models: A survey.
\newblock \emph{arXiv preprint arXiv:2503.15850}, 2025.

\bibitem[Huang et~al.(2023)Huang, Song, Wang, Zhao, Chen, Juefei-Xu, and Ma]{huang2023look}
Yuheng Huang, Jiayang Song, Zhijie Wang, Shengming Zhao, Huaming Chen, Felix Juefei-Xu, and Lei Ma.
\newblock Look before you leap: An exploratory study of uncertainty measurement for large language models.
\newblock \emph{arXiv preprint arXiv:2307.10236}, 2023.

\bibitem[Ling et~al.(2024)Ling, Zhao, Cheng, Liu, Sun, Zhang, Oishi, Osaki, Matsuda, Ji, et~al.]{ling2024uncertainty}
Chen Ling, Xujiang Zhao, Wei Cheng, Yanchi Liu, Yiyou Sun, Xuchao Zhang, Mika Oishi, Takao Osaki, Katsushi Matsuda, Jie Ji, et~al.
\newblock Uncertainty decomposition and quantification for in-context learning of large language models.
\newblock \emph{arXiv e-prints}, pages arXiv--2402, 2024.

\bibitem[Tang et~al.(2024)Tang, Shen, and Kejriwal]{tang2024evaluation}
Zhisheng Tang, Ke~Shen, and Mayank Kejriwal.
\newblock An evaluation of estimative uncertainty in large language models.
\newblock \emph{arXiv preprint arXiv:2405.15185}, 2024.

\bibitem[Chen and Mueller(2023)]{chen2023quantifying}
Jiuhai Chen and Jonas Mueller.
\newblock Quantifying uncertainty in answers from any language model via intrinsic and extrinsic confidence assessment.
\newblock \emph{arXiv preprint arXiv:2308.16175}, 2, 2023.

\bibitem[Yang et~al.(2024)Yang, Yang, Zhang, Hui, Zheng, Yu, Li, Liu, Huang, Wei, et~al.]{yang2024qwen2}
An~Yang, Baosong Yang, Beichen Zhang, Binyuan Hui, Bo~Zheng, Bowen Yu, Chengyuan Li, Dayiheng Liu, Fei Huang, Haoran Wei, et~al.
\newblock Qwen2. 5 technical report.
\newblock \emph{arXiv preprint arXiv:2412.15115}, 2024.

\bibitem[Grattafiori et~al.(2024)Grattafiori, Dubey, Jauhri, Pandey, Kadian, Al-Dahle, Letman, Mathur, Schelten, Vaughan, et~al.]{grattafiori2024llama}
Aaron Grattafiori, Abhimanyu Dubey, Abhinav Jauhri, Abhinav Pandey, Abhishek Kadian, Ahmad Al-Dahle, Aiesha Letman, Akhil Mathur, Alan Schelten, Alex Vaughan, et~al.
\newblock The llama 3 herd of models.
\newblock \emph{arXiv preprint arXiv:2407.21783}, 2024.

\bibitem[Rein et~al.(2024)Rein, Hou, Stickland, Petty, Pang, Dirani, Michael, and Bowman]{rein2024gpqa}
David Rein, Betty~Li Hou, Asa~Cooper Stickland, Jackson Petty, Richard~Yuanzhe Pang, Julien Dirani, Julian Michael, and Samuel~R Bowman.
\newblock Gpqa: A graduate-level google-proof q\&a benchmark.
\newblock In \emph{First Conference on Language Modeling}, 2024.

\bibitem[Wang et~al.(2024)Wang, Ma, Zhang, Ni, Chandra, Guo, Ren, Arulraj, He, Jiang, et~al.]{wang2024mmlu}
Yubo Wang, Xueguang Ma, Ge~Zhang, Yuansheng Ni, Abhranil Chandra, Shiguang Guo, Weiming Ren, Aaran Arulraj, Xuan He, Ziyan Jiang, et~al.
\newblock Mmlu-pro: A more robust and challenging multi-task language understanding benchmark.
\newblock In \emph{The Thirty-eight Conference on Neural Information Processing Systems Datasets and Benchmarks Track}, 2024.

\bibitem[Du et~al.(2025)Du, Yao, Ma, Wang, Zheng, Zhu, Liu, Liang, Jin, Wei, et~al.]{du2025supergpqa}
Xinrun Du, Yifan Yao, Kaijing Ma, Bingli Wang, Tianyu Zheng, King Zhu, Minghao Liu, Yiming Liang, Xiaolong Jin, Zhenlin Wei, et~al.
\newblock Supergpqa: Scaling llm evaluation across 285 graduate disciplines.
\newblock \emph{arXiv preprint arXiv:2502.14739}, 2025.

\bibitem[Balabanov and Linander(2024)]{balabanov2024uncertainty}
Oleksandr Balabanov and Hampus Linander.
\newblock Uncertainty quantification in fine-tuned llms using lora ensembles.
\newblock \emph{arXiv preprint arXiv:2402.12264}, 2024.

\bibitem[Shorinwa et~al.(2024)Shorinwa, Mei, Lidard, Ren, and Majumdar]{shorinwa2024survey}
Ola Shorinwa, Zhiting Mei, Justin Lidard, Allen~Z Ren, and Anirudha Majumdar.
\newblock A survey on uncertainty quantification of large language models: Taxonomy, open research challenges, and future directions.
\newblock \emph{arXiv preprint arXiv:2412.05563}, 2024.

\bibitem[Azaria and Mitchell(2023)]{azaria2023internal}
Amos Azaria and Tom Mitchell.
\newblock The internal state of an llm knows when it's lying.
\newblock \emph{arXiv preprint arXiv:2304.13734}, 2023.

\bibitem[Zhang et~al.(2024{\natexlab{b}})Zhang, Liu, Basaldella, and Collier]{zhang2024luq}
Caiqi Zhang, Fangyu Liu, Marco Basaldella, and Nigel Collier.
\newblock Luq: Long-text uncertainty quantification for llms.
\newblock \emph{arXiv preprint arXiv:2403.20279}, 2024{\natexlab{b}}.

\bibitem[Fadeeva et~al.(2023)Fadeeva, Vashurin, Tsvigun, Vazhentsev, Petrakov, Fedyanin, Vasilev, Goncharova, Panchenko, Panov, et~al.]{fadeeva2023lm}
Ekaterina Fadeeva, Roman Vashurin, Akim Tsvigun, Artem Vazhentsev, Sergey Petrakov, Kirill Fedyanin, Daniil Vasilev, Elizaveta Goncharova, Alexander Panchenko, Maxim Panov, et~al.
\newblock Lm-polygraph: Uncertainty estimation for language models.
\newblock \emph{arXiv preprint arXiv:2311.07383}, 2023.

\bibitem[Manakul et~al.(2023)Manakul, Liusie, and Gales]{manakul2023selfcheckgpt}
Potsawee Manakul, Adian Liusie, and Mark~JF Gales.
\newblock Selfcheckgpt: Zero-resource black-box hallucination detection for generative large language models.
\newblock \emph{arXiv preprint arXiv:2303.08896}, 2023.

\bibitem[Fadeeva et~al.(2024)Fadeeva, Rubashevskii, Shelmanov, Petrakov, Li, Mubarak, Tsymbalov, Kuzmin, Panchenko, Baldwin, et~al.]{fadeeva2024fact}
Ekaterina Fadeeva, Aleksandr Rubashevskii, Artem Shelmanov, Sergey Petrakov, Haonan Li, Hamdy Mubarak, Evgenii Tsymbalov, Gleb Kuzmin, Alexander Panchenko, Timothy Baldwin, et~al.
\newblock Fact-checking the output of large language models via token-level uncertainty quantification.
\newblock \emph{arXiv preprint arXiv:2403.04696}, 2024.

\bibitem[Liu et~al.(2019)Liu, Ott, Goyal, Du, Joshi, Chen, Levy, Lewis, Zettlemoyer, and Stoyanov]{liu2019roberta}
Yinhan Liu, Myle Ott, Naman Goyal, Jingfei Du, Mandar Joshi, Danqi Chen, Omer Levy, Mike Lewis, Luke Zettlemoyer, and Veselin Stoyanov.
\newblock Roberta: A robustly optimized bert pretraining approach.
\newblock \emph{arXiv preprint arXiv:1907.11692}, 2019.

\bibitem[Reimers and Gurevych(2019)]{reimers2019sentence}
Nils Reimers and Iryna Gurevych.
\newblock Sentence-bert: Sentence embeddings using siamese bert-networks.
\newblock \emph{arXiv preprint arXiv:1908.10084}, 2019.

\bibitem[Zhang et~al.(2019)Zhang, Kishore, Wu, Weinberger, and Artzi]{zhang2019bertscore}
Tianyi Zhang, Varsha Kishore, Felix Wu, Kilian~Q Weinberger, and Yoav Artzi.
\newblock Bertscore: Evaluating text generation with bert.
\newblock \emph{arXiv preprint arXiv:1904.09675}, 2019.

\bibitem[Hou et~al.(2023)Hou, Liu, Qian, Andreas, Chang, and Zhang]{hou2023decomposing}
Bairu Hou, Yujian Liu, Kaizhi Qian, Jacob Andreas, Shiyu Chang, and Yang Zhang.
\newblock Decomposing uncertainty for large language models through input clarification ensembling.
\newblock \emph{arXiv preprint arXiv:2311.08718}, 2023.

\bibitem[Min et~al.(2020)Min, Michael, Hajishirzi, and Zettlemoyer]{min2020ambigqa}
Sewon Min, Julian Michael, Hannaneh Hajishirzi, and Luke Zettlemoyer.
\newblock Ambigqa: Answering ambiguous open-domain questions.
\newblock \emph{arXiv preprint arXiv:2004.10645}, 2020.

\bibitem[Kuhn et~al.(2022)Kuhn, Gal, and Farquhar]{kuhn2022clam}
Lorenz Kuhn, Yarin Gal, and Sebastian Farquhar.
\newblock Clam: Selective clarification for ambiguous questions with generative language models.
\newblock \emph{arXiv preprint arXiv:2212.07769}, 2022.

\bibitem[Hadi et~al.(2023)Hadi, Qureshi, Shah, Irfan, Zafar, Shaikh, Akhtar, Wu, Mirjalili, et~al.]{hadi2023survey}
Muhammad~Usman Hadi, Rizwan Qureshi, Abbas Shah, Muhammad Irfan, Anas Zafar, Muhammad~Bilal Shaikh, Naveed Akhtar, Jia Wu, Seyedali Mirjalili, et~al.
\newblock A survey on large language models: Applications, challenges, limitations, and practical usage.
\newblock \emph{Authorea Preprints}, 2023.

\bibitem[Li et~al.(2024{\natexlab{b}})Li, Wen, Wang, Li, Yuan, Liu, Liu, Xu, Wang, Sun, et~al.]{li2024personal}
Yuanchun Li, Hao Wen, Weijun Wang, Xiangyu Li, Yizhen Yuan, Guohong Liu, Jiacheng Liu, Wenxing Xu, Xiang Wang, Yi~Sun, et~al.
\newblock Personal llm agents: Insights and survey about the capability, efficiency and security.
\newblock \emph{arXiv preprint arXiv:2401.05459}, 2024{\natexlab{b}}.

\bibitem[Jablonka et~al.(2024)Jablonka, Schwaller, Ortega-Guerrero, and Smit]{jablonka2024leveraging}
Kevin~Maik Jablonka, Philippe Schwaller, Andres Ortega-Guerrero, and Berend Smit.
\newblock Leveraging large language models for predictive chemistry.
\newblock \emph{Nature Machine Intelligence}, 6\penalty0 (2):\penalty0 161--169, 2024.

\bibitem[Li et~al.(2024{\natexlab{c}})Li, Liu, Fan, Wei, Liu, Tang, and Li]{li2024empowering}
Jiatong Li, Yunqing Liu, Wenqi Fan, Xiao-Yong Wei, Hui Liu, Jiliang Tang, and Qing Li.
\newblock Empowering molecule discovery for molecule-caption translation with large language models: A chatgpt perspective.
\newblock \emph{IEEE transactions on knowledge and data engineering}, 2024{\natexlab{c}}.

\bibitem[Boiko et~al.(2023)Boiko, MacKnight, Kline, and Gomes]{boiko2023autonomous}
Daniil~A Boiko, Robert MacKnight, Ben Kline, and Gabe Gomes.
\newblock Autonomous chemical research with large language models.
\newblock \emph{Nature}, 624\penalty0 (7992):\penalty0 570--578, 2023.

\bibitem[Zhao et~al.(2024)Zhao, Ma, Chen, Sun, Li, Xia, Chen, Xu, Zhu, Zhu, et~al.]{zhao2024chemdfm}
Zihan Zhao, Da~Ma, Lu~Chen, Liangtai Sun, Zihao Li, Yi~Xia, Bo~Chen, Hongshen Xu, Zichen Zhu, Su~Zhu, et~al.
\newblock Chemdfm: A large language foundation model for chemistry.
\newblock \emph{arXiv preprint arXiv:2401.14818}, 2024.

\bibitem[Huang et~al.(2024{\natexlab{b}})Huang, Zhang, He, Zhi, Wang, Li, Xu, Liu, Liang, Li, et~al.]{huang2024chemeval}
Yuqing Huang, Rongyang Zhang, Xuesong He, Xuyang Zhi, Hao Wang, Xin Li, Feiyang Xu, Deguang Liu, Huadong Liang, Yi~Li, et~al.
\newblock Chemeval: A comprehensive multi-level chemical evaluation for large language models.
\newblock \emph{arXiv preprint arXiv:2409.13989}, 2024{\natexlab{b}}.

\bibitem[Malinin and Gales(2020)]{malinin2020uncertainty}
Andrey Malinin and Mark Gales.
\newblock Uncertainty estimation in autoregressive structured prediction.
\newblock \emph{arXiv preprint arXiv:2002.07650}, 2020.

\bibitem[Bakman et~al.(2024)Bakman, Yaldiz, Buyukates, Tao, Dimitriadis, and Avestimehr]{bakman2024mars}
Yavuz~Faruk Bakman, Duygu~Nur Yaldiz, Baturalp Buyukates, Chenyang Tao, Dimitrios Dimitriadis, and Salman Avestimehr.
\newblock Mars: Meaning-aware response scoring for uncertainty estimation in generative llms.
\newblock In \emph{Proceedings of the 62nd Annual Meeting of the Association for Computational Linguistics (Volume 1: Long Papers)}, pages 7752--7767, 2024.

\bibitem[Duan et~al.(2023)Duan, Cheng, Wang, Zavalny, Wang, Xu, Kailkhura, and Xu]{duan2023shifting}
Jinhao Duan, Hao Cheng, Shiqi Wang, Alex Zavalny, Chenan Wang, Renjing Xu, Bhavya Kailkhura, and Kaidi Xu.
\newblock Shifting attention to relevance: Towards the predictive uncertainty quantification of free-form large language models.
\newblock \emph{arXiv preprint arXiv:2307.01379}, 2023.

\end{thebibliography}

%%% Uncomment this section and comment out the \bibliography{references} line above to use inline references.
% \begin{thebibliography}{1}

% 	\bibitem{kour2014real}
% 	George Kour and Raid Saabne.
% 	\newblock Real-time segmentation of on-line handwritten arabic script.
% 	\newblock In {\em Frontiers in Handwriting Recognition (ICFHR), 2014 14th
% 			International Conference on}, pages 417--422. IEEE, 2014.

% 	\bibitem{kour2014fast}
% 	George Kour and Raid Saabne.
% 	\newblock Fast classification of handwritten on-line arabic characters.
% 	\newblock In {\em Soft Computing and Pattern Recognition (SoCPaR), 2014 6th
% 			International Conference of}, pages 312--318. IEEE, 2014.

% 	\bibitem{hadash2018estimate}
% 	Guy Hadash, Einat Kermany, Boaz Carmeli, Ofer Lavi, George Kour, and Alon
% 	Jacovi.
% 	\newblock Estimate and replace: A novel approach to integrating deep neural
% 	networks with existing applications.
% 	\newblock {\em arXiv preprint arXiv:1804.09028}, 2018.

% \end{thebibliography}

\end{document}